\documentclass[sigconf]{acmart}
\AtBeginDocument{%
  \providecommand\BibTeX{{%
    \normalfont B\kern-0.5em{\scshape i\kern-0.25em b}\kern-0.8em\TeX}}}
\DeclareUnicodeCharacter{0301}{\'{e}} 
\usepackage[caption=false,font=footnotesize]{subfig}

\begin{document}
\title{Recent Advances in Bayesian Optimization}


\author{Xilu Wang}
\affiliation{%
  \institution{Department of Computer Science, University of Surrey}
  \city{Guildford}
  \country{United Kingdom}
  \postcode{GU2 7XH}}
\email{xilu.wang@surrey.ac.uk}

\author{Yaochu Jin}
\affiliation{%
  \institution{Faculty of Technology, Bielefeld University}
  \city{33619 Bielefeld}
  \country{Germany}}
\email{yaochu.jin@uni-bielefeld.de}  

\author{Sebastian Schmitt}
\author{Markus Olhofer}
\affiliation{%
  \institution{Honda Research Institute Europe GmbH}
  \streetaddress{Carl-Legien-Strasse 30}
  \city{63073 Offenbach/Main}
  \country{Germany}}
\email{{sebastian.schmitt;markus.olhofer}@honda-ri.de}

\newcommand{\seb}[1]{\textcolor[rgb]{0,0.6,1.}{#1}}
\newcommand{\sebC}[1]{\textcolor[rgb]{0,0.6,0.5}{Comment: #1}}

\renewcommand{\shortauthors}{Wang et al.}

\begin{abstract}
 Bayesian optimization has emerged at the forefront of expensive black-box optimization due to its data efficiency. Recent years have witnessed a proliferation of studies on the development of new Bayesian optimization algorithms and their applications. Hence, this paper attempts to provide a comprehensive and updated survey of recent advances in Bayesian optimization and identify interesting open problems. We categorize the existing work on Bayesian optimization into nine main groups according to the motivations and focus of the proposed algorithms. For each category, we present the main advances with respect to the construction of surrogate models and adaptation of the acquisition functions. Finally, we discuss the open questions and suggest promising future research directions, in particular with regard to heterogeneity, privacy preservation, and fairness in distributed and federated optimization systems. 
\end{abstract}



\maketitle
\section{Introduction}
Optimization problems are pervasive in scientific and industrial fields, such as artificial intelligence, data mining, bioinformatics, software engineering, scheduling, manufacturing, and economics. Among them, many applications require to optimize objective functions that are noisy and expensive to evaluate, or do not have closed-form expressions, let alone gradient information. For such problems, metaheuristics such as evolutionary algorithms that rely on function values only are very popular. However, these algorithms usually require a large number of function evaluations. By contrast, Bayesian optimization has emerged as a mainstream to tackle these difficulties due to its high data efficiency, thanks to its ability to incorporate prior beliefs about the problem to help guide the sampling of new data, and to achieve a good balance between exploration and exploitation in the search.

Consider the maximization of an unknown function $f$ that is expensive to evaluate, which can be formulated as follows:
\begin{equation}
x^{*}=\arg \max _{x \in \mathcal{X}} f(x)
\end{equation}
where $\mathcal{X}$ denotes the search/decision space of interest and $x^{*}$ is the global maximum. In principle, Bayesian optimization constructs a probabilistic model (also known as a surrogate model) that defines a distribution over the objective function, and then subsequently refines this model once new data is sampled. Specifically, Bayesian optimization first specifies a prior distribution over the function, which represents our belief about the objective function. Then, conditioned on the observed data and the prior, the posterior can be calculated using the Bayes rule, which quantifies our updated belief about the unknown objective function. As a result, the next sample can be identified by leveraging the posterior. This is achieved by optimizing some auxiliary functions, called acquisition functions in Bayesian optimization. 

The origin of Bayesian optimization can be dated back to the work by Harold Kushner \cite{kushner1964new}, where Wiener processes were adopted for unconstrained one-dimensional optimization problems and the probability of improvement is maximized to select the next sample. Mockus \cite{movckus1975bayesian} developed a new acquisition function, called expectation of improvement (EI), which was further used in \cite{zilinskas1978optimization}. Stuckman \cite{stuckman1988global}, Perttunen \cite{perttunen1990rank} and Elder \cite{elder1992global} extended Kushner's work to high-dimensional problems.  
Bayesian optimization was made popular in engineering after Jones \emph{et al.} \cite{jones1998efficient} introduced Efficient Global Optimization (EGO). In EGO, a Kriging model, called Design and Analysis of Computer Experiments (DACE) stochastic process model \cite{sacks1989design}, is adopted to provide best linear unbiased predictions of the objective, which is achieved by minimizing the Mean Squared Error of the predictor \cite{kleijnen2009kriging}. In Bayesian optimization, by contrast, a Gaussian process is adopted as the surrogate model, which is fit by maximizing the likelihood. Hence, the original formulation of Kriging is different from the Gaussian process \cite{cressie1990origins}. More recently, various variants of Kriging have been developed \cite{van2003kriging,huang2006global} by accounting for constraints and noise in the optimization. As a result, Kriging models in spatial statistics are equivalent to Gaussian processes in Bayesian optimization in some papers, therefore the two terms will be used interchangeably in the rest of this paper.
The past decades have witnessed a rapid development of Bayesian optimization in many real-world problems, including materials design and discovery \cite{frazier2016bayesian}, sensor network \cite{garnett2010bayesian}, financial industry \cite{gonzalvez2019financial}, and experimental design \cite{lorenz2019efficiently}. More recently, Bayesian optimization became popular in machine learning, including reinforcement learning \cite{turchetta2020robust}, hyperparameter tuning \cite{wu2020practical}, and neural architecture search \cite{kandasamy2018neural}. 

\subsection{Related Surveys}
There are already a few comprehensive surveys and tutorials on methodological and practical aspects of Bayesian optimization, each with a specific focus.  Sasena \cite{survey1sasena2002} gave a review of early work on Kriging and its extension to constrained optimization. A tutorial on Bayesian optimization with Gaussian processes was given in \cite{survey2brochu2010}, focusing on extending Bayesian optimization to active user modelling in preference galleries and hierarchical control problems. Shahriari \emph{et al.} \cite{survey4shahriari2016} presented a comprehensive review of the fundamentals of Bayesian optimization, elaborating on the statistical modeling and popular acquisition functions. In addition, Frazier \cite{survey5frazier2018} discussed some recent advances in Bayesian optimization, in particular in multi-fidelity optimization and constrained optimization. 

However, none of the above review papers provides a comprehensive coverage of abundant extensions of Bayesian optimization. Moreover, many new advances in Bayesian optimization have been published since \cite{survey4shahriari2016}. Hence, an updated and comprehensive survey of this dynamic research field will be beneficial for researchers and practitioners. 

\subsection{Contributions and Organization}
\begin{figure}[ht]
\centering

\includegraphics[width=0.75\columnwidth]{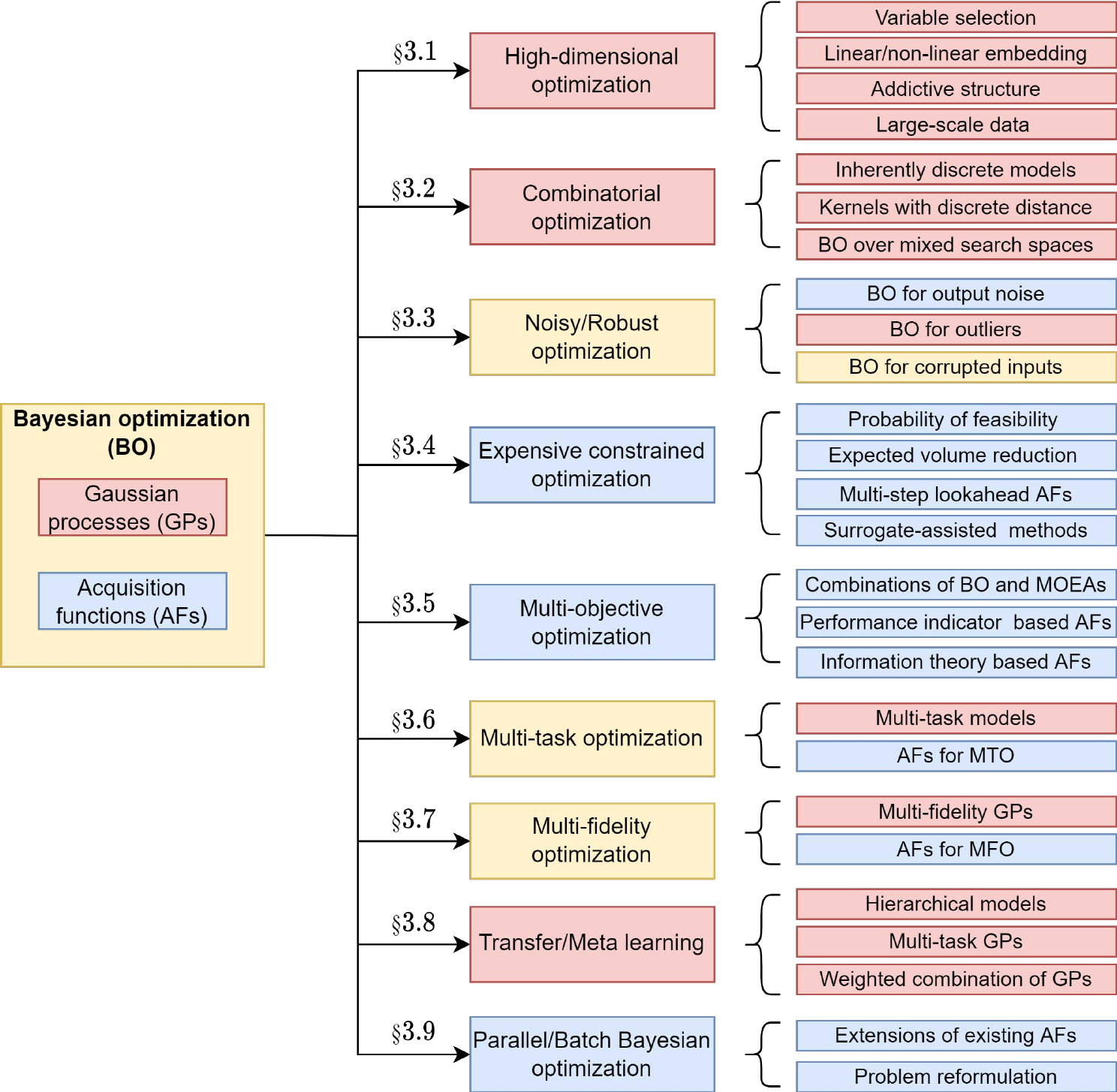}
\caption{Taxonomy of Bayesian optimization algorithms. In the diagram, BO stands for Bayesian optimization, GP for Gaussian process, AF for acquisition function, MOEA for multi-objective evolutionary algorithm, MFO for multi-fidelity optimization, and MTO for multi-task optimization.}
\label{BO_CATEGORITION}
\end{figure}
This paper starts with a brief introduction to the fundamentals of Bayesian optimization in Section \ref{Fundamentals of Bayesian Optimization}, including Gaussian processes and commonly used acquisition functions. Section \ref{Recent Advances} provides a comprehensive review of the state-of-the-art, where a taxonomy of existing work on Bayesian optimization is proposed to offer a clear structure of the large body of research reported in the literature, as illustrated in Fig. \ref{BO_CATEGORITION}. In this taxonomy, existing Bayesian optimization algorithms are divided into nine groups according to the nature of the optimization problems. We further introduce a color-coding scheme to highlight the focuses of each group, where red, blue and yellow blocks indicate, respectively, a focus on acquisition functions, surrogates, or both. Finally, this survey explores a few emerging topics in Bayesian optimization, including Bayesian dynamic optimization, distributed and federated Bayesian optimization, and heterogeneity and fairness in optimization. 
\section{Fundamentals of Bayesian Optimization} 
\label{Fundamentals of Bayesian Optimization}
Gaussian processes and acquisition functions are two main components of Bayesian optimization, which are introduced in the following. 
\subsection{Gaussian Process}
 Gaussian process (GP) is the most widely used probabilistic surrogate model for approximating the true objective function in Bayesian optimization. GP is characterized by a prior mean function $\mu(\cdot)$ and a covariance function $\kappa(\cdot, \cdot)$ \cite{rasmussen2003gaussian}. 
Consider a finite collection of data pairs $\mathcal{D}_{n}=(\mathbf{X},\mathbf{y})$ of the unknown function $y=f(\mathbf{X})+\epsilon $ with  noise $\epsilon \sim \mathcal{N}\left(0, \sigma_{\epsilon}^{2}\right)$, where $\mathbf{X}=[\mathbf{x}_{1}, \mathbf{x}_{2}, \cdots, \mathbf{x}_{n}]^{T}$ is the input and $\mathbf{y} =[y_{1}, y_{2}, \cdots, y_{n}]^{T}$ is the output resulting from the true objective evaluations, and $n$ is the number of samples. The Gaussian process model assumes that the observed data are drawn from a multivariate Gaussian distribution. Therefore, for a new data point $\mathbf{x}$, the joint distribution of the observed outputs $\mathbf{y}$ and the predicted output $y$ are
\begin{equation}
\left[\begin{array}{l}
\mathbf{y} \\
y
\end{array}\right] \sim \mathcal{N}\left(\mathbf{0},\left[\begin{array}{ll}
K(\mathbf{X}, \mathbf{X}) +\sigma_{\epsilon}^{2} \mathbf{I}& K\left(\mathbf{X}, \mathbf{x}\right) \\
K\left(\mathbf{X}, \mathbf{x}\right)^{T} & \kappa \left(\mathbf{x}, \mathbf{x}\right)
\end{array}\right]\right)
\end{equation}
where $T$ denotes matrix transposition, $K(\mathbf{X}, \mathbf{X})=[\kappa(\mathbf{x}_{i}, \mathbf{x}_{j})]_{\mathbf{x}_{i}, \mathbf{x}_{j}\in \mathbf{X}}$ denotes an $n\times n$ correlation matrix, and $K(\boldsymbol{X}, \mathbf{x})=[\kappa(\mathbf{x}_{i}, \mathbf{x})]_{\mathbf{x}_{i}\in \mathbf{X}}$ denotes a correlation vector evaluated at all pairs of training and test points. As described in \cite{rasmussen2003gaussian}, the conditional distribution $p(y \mid \mathbf{x}, \mathbf{X}, \mathbf{y}) \sim \mathcal{N}(\mu(\mathbf{x}),\sigma^2 (\mathbf{x}))$ is then a multivariate Gaussian distribution, where the mean and variance of the predicted output $y$ can be estimated as
\begin{equation}
\begin{aligned}
\mu(\mathbf{x})&=K\left(\mathbf{x}, \mathbf{X}\right) \left(K(\mathbf{X}, \mathbf{X}\right)+\sigma_{\epsilon}^{2} \mathbf{I})^{-1} \mathbf{y}\\
\sigma^2 (\mathbf{x})&=\kappa \left(\mathbf{x}, \mathbf{x}\right)-K\left(\mathbf{X}, \mathbf{x}\right)^{T}(K(\mathbf{X}, \mathbf{X})+\sigma_{\epsilon}^{2} \mathbf{I})^{-1} K\left(\mathbf{X}, \mathbf{x}\right).
\end{aligned}
\end{equation}

Commonly used kernel functions are the squared exponential (Gaussian) kernel and the Mat\'{a}ern kernel \cite{survey5frazier2018}, where hyperparameters, such as length scale, signal variance, and noise variance need to be specified. Typically, the optimal hyperparameters are inferred by maximizing the log marginal likelihood,
\begin{equation}
\log p(\mathbf{y} \mid \mathbf{X}, \mathbf{\theta})=-\frac{1}{2} \mathbf{y}^{T} \mathbf{K}_{y}^{-1} \mathbf{y}-\frac{1}{2} \log \left|\mathbf{K}_{y}\right|-\frac{n}{2} \log 2 \pi
\label{MarginalLikelihood}
\end{equation}
where $\mathbf{K}_{y}=K(\mathbf{X}, \mathbf{X})+\sigma_{\epsilon}^{2} \mathbf{I}$. 

\subsection{Acquisition Function}
\begin{figure}[!htbp]
\centerline{
\subfloat[Iteration=1]{\includegraphics[width=1.8in]{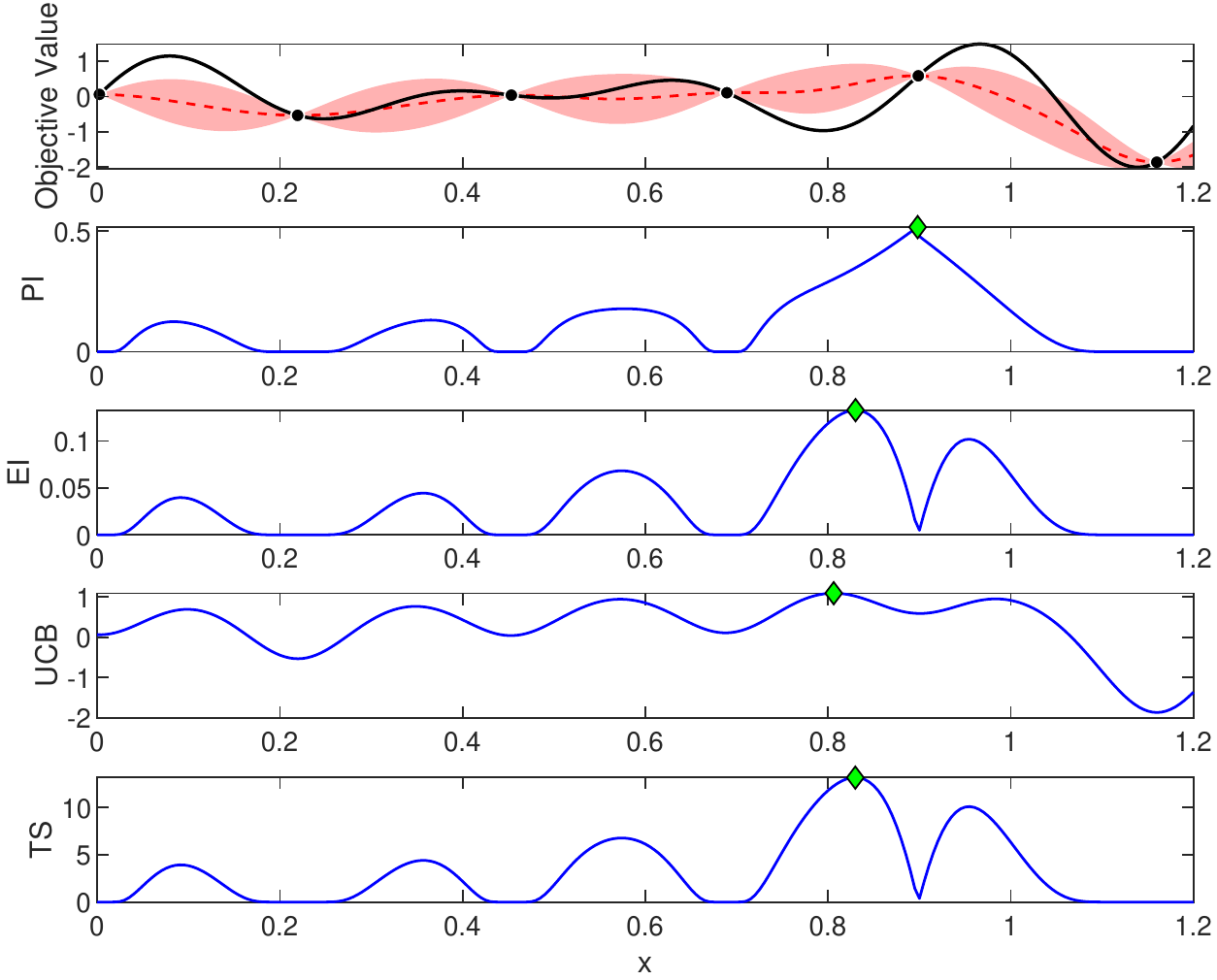}%
\hfil
}
\subfloat[Iteration=2]{\includegraphics[width=1.8in]{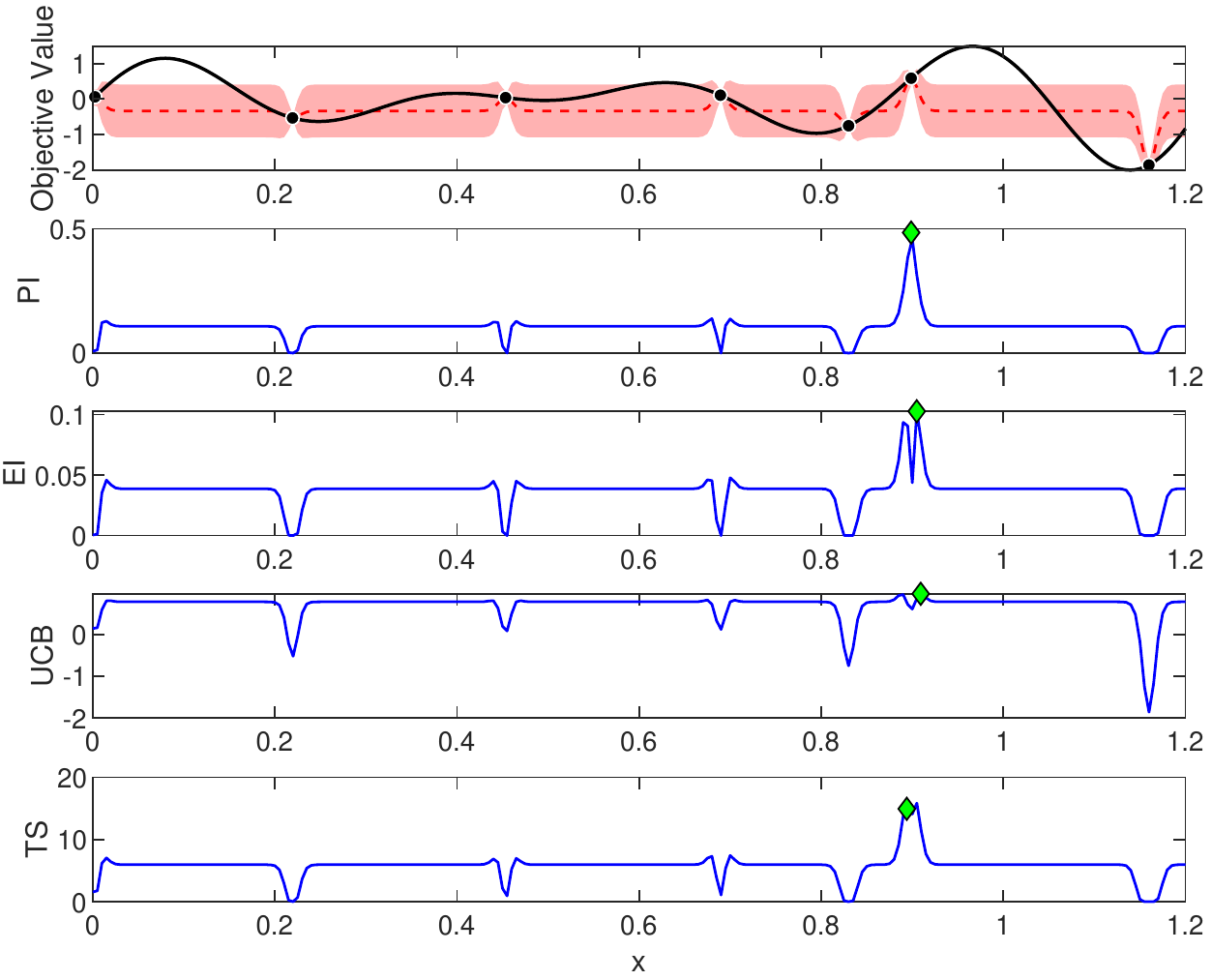}%
\hfil
}
}
\centerline{
\subfloat[Iteration=3]{\includegraphics[width=1.8in]{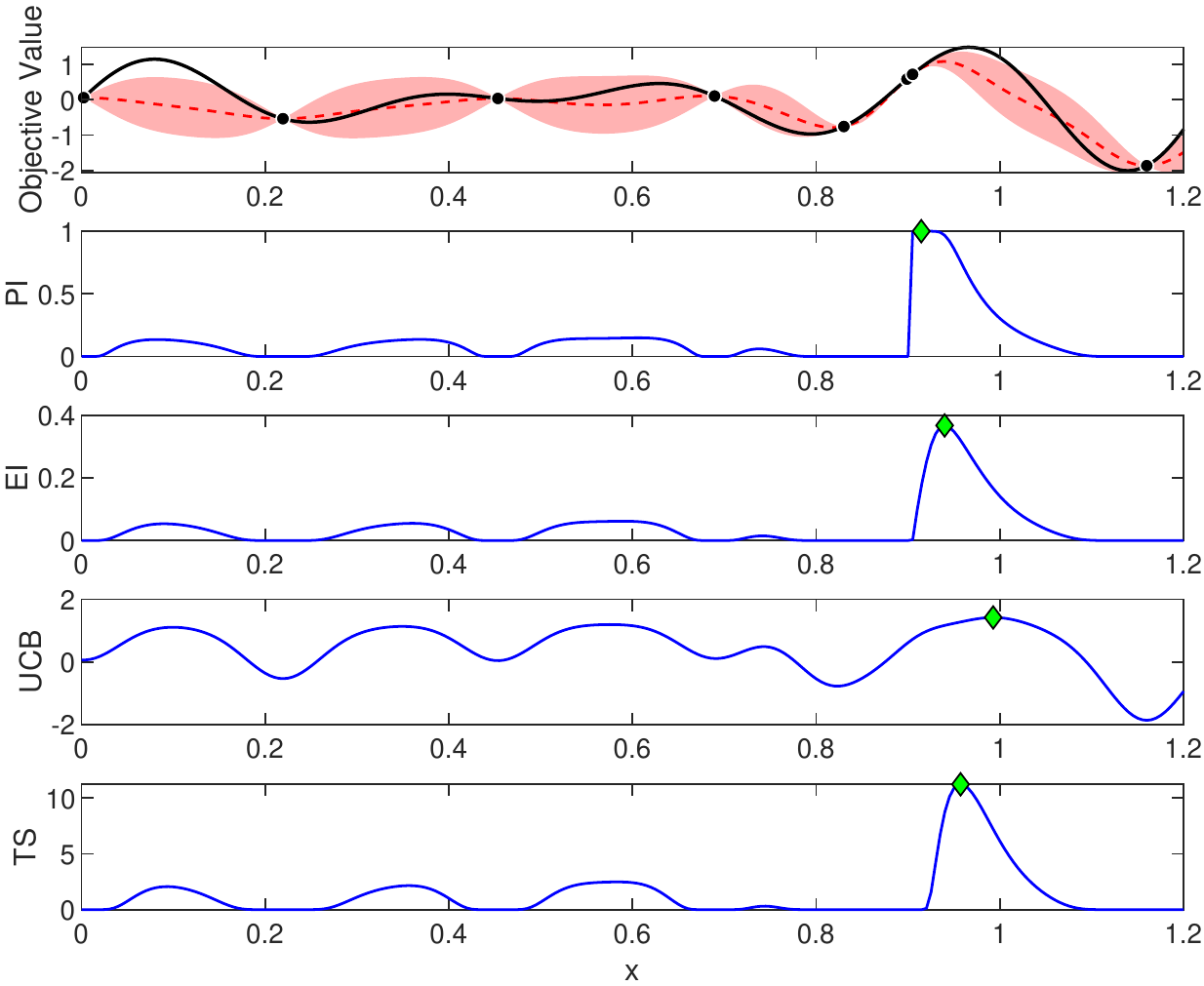}%
\hfil
}
\subfloat[Iteration=4]{\includegraphics[width=1.8in]{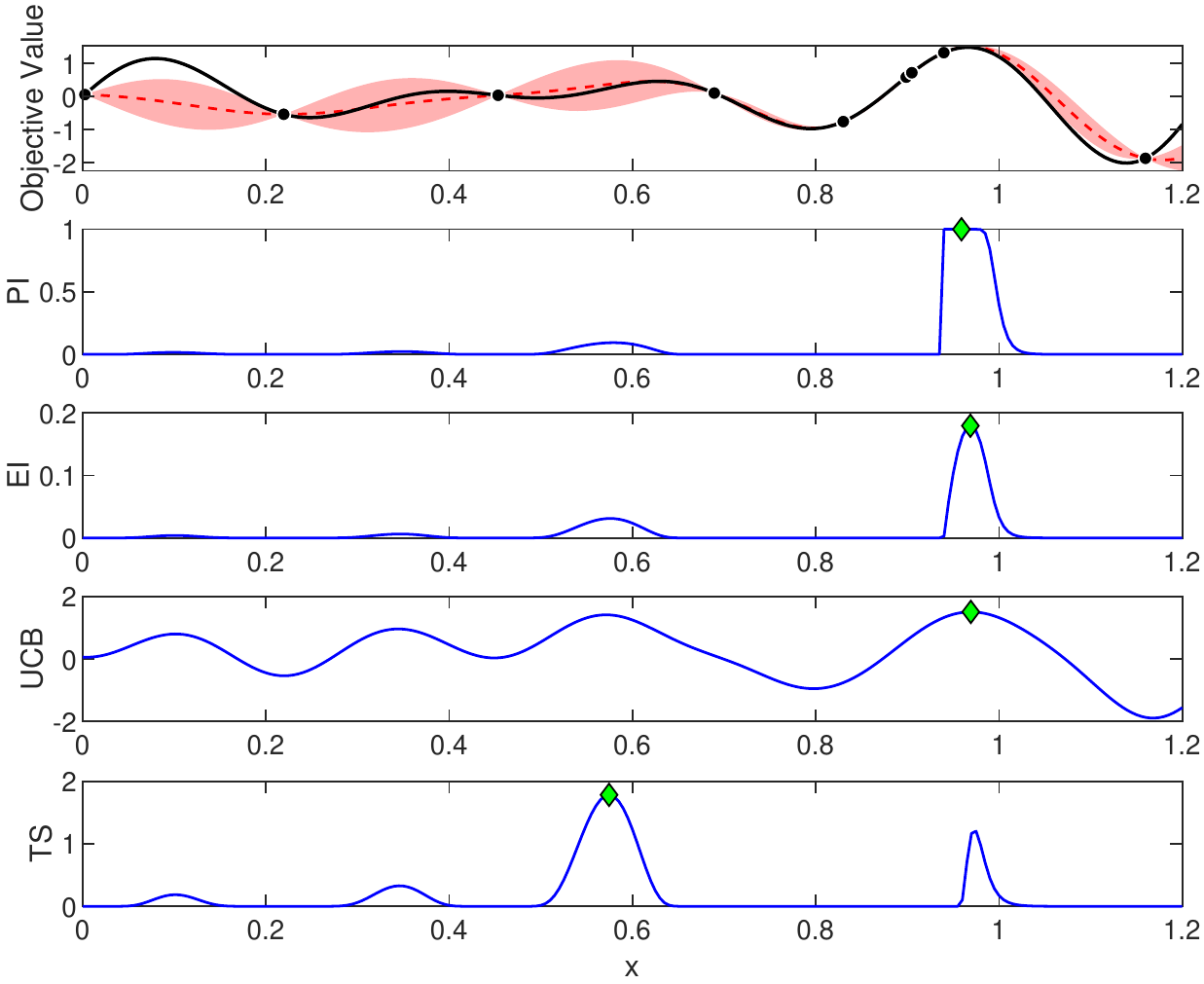}%
\hfil
}
}
\caption{Illustration of Bayesian optimization with acquisition functions, maximizing a 1D black-box function (solid black line) with four iterations. In each sub-figure, the top row shows the observations obtained so far, the predicted mean values (dotted red line) and variance (red shaded region) provided by the Gaussian process. The lower four rows show the four acquisition functions (blue lines), probability of improvement, expected improvement, upper confidence bound, and Thompson sampling (from top to bottom), and the corresponding new samples (green diamonds). Note that at each iteration the new sample identified by expected improvement is adopted to update the Gaussian process.}
BO starts by sampling a training set from the black-box function, by which a Gaussian process is constructed. At each iteration, an acquisition function is evaluated based the GP and optimised to identify where to sample next (green point) from the true objective function. The new sample is added into the training set to update the model. This procedure is repeated until until the termination condition is met.
\label{BO_GPandAF}
\end{figure}

Acquisition functions (AFs) are the utility functions that guide the search to reach the optimum of the objective function by identifying where to sample next, which is crucial in Bayesian optimization. The guiding principle behind AFs is to strike a balance between exploration and exploitation, which is achieved by querying samples from both known high-fitness-value regions and regions that have not been sufficiently explored so far. 
While the top row in each panel shows a Gaussian process, the lower four rows illustrate four commonly used AFs. In the following, we briefly revisit the commonly used AFs. 

Without loss of generality, we consider a maximization problem. Let $f^*$ denote the optimum obtained so far, 
and $\Phi(\cdot)$ and $\phi(\cdot)$ denote the normal cumulative distribution function (CDF), and probability density function (PDF) of the standard normal random variable, respectively. The earliest acquisition function is to maximize the \emph{probability of improvement} (PI) \cite{kushner1964new} over the current best value $f^*$, formulated as 
\begin{equation}
\mathrm{PI}(\mathbf{x}) =P\left(f(\mathbf{x}) \geq f^*\right) 
=\Phi\left(\frac{\mu(\mathbf{x})-f^*}{\sigma(\mathbf{x})}\right),
\end{equation}
where $P$ is the probability for finding a better objective function value at position $\mathbf{x}$ than the currently best value $f^*$, which, for a Gaussian process, is given by the Gaussian cumulative distribution function (CDF) $\Phi(x)=\tfrac{1}{\sqrt{2\pi}}\int_{-\infty}^x\exp(-t^2/2)dt$.

Alternatively, \emph{expected improvement} (EI) \cite{movckus1975bayesian} calculates the expected improvement with respect to $f^*$,
\begin{equation}
\begin{aligned}
\operatorname{EI}(\mathbf{x}) &=\mathbb{E}\left[\max \left(0, f^*-f(\mathbf{x})\right) \right] \\
&=\left(f^*-\mu(\mathbf{x})\right) \Phi\left(\frac{f^*-\mu(\mathbf{x})}{\sigma(\mathbf{x})}\right)+\sigma(\mathbf{x}) \phi\left(\frac{f^*-\mu(\mathbf{x})}{\sigma(\mathbf{x})}\right),
\end{aligned}
\label{EI}
\end{equation}
where $\mathbb{E}$ denotes the expectation value, $\Phi$ and $\phi$ are the Gaussian CDF and PDF, respectively.
EI dates back to 1975 \cite{movckus1975bayesian} and was popularized by Jones \emph{et al.} \cite{jones1998efficient}. A wealth of research has been dedicated to the development of EI in various applications, including parallel optimization, multi-objective optimization, constrained optimization, noisy optimization, multi-fidelity optimization, and high-dimensional optimization. 

Interested readers are referred to \cite{zhan2020expected} for a comprehensive review of many variants of EI. Note, however, that EI tends to explore around the initial best point before the algorithm begins to search more globally, as only points that are close to the current best point have high EI values. 

An idea closely related to EI is \emph{Knowledge Gradient} (KG) \cite{frazier2008knowledge}, maximizing the expected incremental value of a measurement; however, it does not depend on the optimum obtained so far. Let $\mu_{n}$ denote the mean of the posterior distribution after $n$ samples, and a new posterior distribution with posterior mean $\mu_{n+1}$ will be generated if we take one more sample. Hence, the KG is formulated as
\begin{equation}
\operatorname{KG}(\mathbf{x})=\mathbb{E}_{n}\left[\max(\mu_{n+1})-\max(\mu_{n}) \right]
\end{equation}
where $\mathbb{E}_{n}[\cdot]:=\mathbb{E}\left[\cdot \mid \mathbf{X}, \mathbf{y}\right]$ indicates the conditional expectation with respect to what is known after the first $n$ measurements. 

The confidence bound criteria, \emph{upper confidence bound} (UCB) for maximization problems and \emph{lower confidence bound} (LCB) for minimization problems, are designed to achieve optimal regret in the multi-armed bandit community by combining the uncertainty and the expected reward \cite{srinivas2009gaussian}. 
The UCB is calculated as
\begin{equation}
\operatorname{UCB}\left(\mathbf{x}\right)=\mu(\mathbf{x})+\beta \sigma(\mathbf{x}),
\end{equation}
where $\beta>0$ is a parameter to navigate the exploitation-exploration trade-off (LCB has a minus sign in front of the $\beta$ term). A recent work \cite{de2021greed} presents $\epsilon$-greedy acquisition functions, where the location with the most promising mean prediction is usually selected, while a multi-objective optimiser is used to generate the Pareto-optimal solutions in the remaining cases. 

Another promising acquisition function for multi-armed bandit problems is \emph{Thompson sampling} (TS) \cite{agrawal2012analysis}. TS randomly draws each arm sampled from the posterior distribution, and then plays the arm with the highest simulated reward \cite{russo2017tutorial}. More recently, TS has seen a surge of interests spurred by the fact that TS can be fully parallelized and distributed \cite{hernandez2017parallel,garcia2019fully,dai2020federated,kandasamy2018parallelised}.

A more recent development is the entropy-based AFs motivated by information theory, which can be further divided into input-entropy-based and output-entropy-based AFs. The former maximizes information about the location $\mathbf{x^*}$ of the global optimum where the information about $\mathbf{x^*}$ is measured by the negative differential entropy of the probability of the location of the global optimum, $p\left(\mathbf{x}^{*} \mid \mathcal{D}_{n}\right)$ \cite{hennig2012entropy,hernandez2014predictive}. Hennig and Schuler \cite{hennig2012entropy} proposed \emph{entropy search} (ES) using mutual information $I\left(\{\mathbf{x}, y\} ; \mathbf{x}^{*} \mid \mathcal{D}_{n}\right)$,
\begin{equation}
\begin{aligned}
&\operatorname{ES}=I\left(\{\mathbf{x}, y\} ; \mathbf{x}^{*} \mid \mathcal{D}_{n}\right) \\
&\quad=\mathrm{H}\left[p\left(\mathbf{x}^{*} \mid
\mathcal{D}_{n}\right)\right]-\mathbb{E}_{p
\left(y \mid \mathcal{D}_{n},
\mathbf{x}\right)}\left[\mathrm{H}\left
[p\left(\mathbf{x}^{*} 
\mid \mathcal{D}_{n} 
\cup\{(\mathbf{x}, y)\}\right)\right]\right],
\label{ES}
\end{aligned}
\end{equation}
where $\mathrm{H}[p(\mathbf{x})]=-\int p(\mathbf{x}) \log p(\mathbf{x}) d \mathbf{x}$ denotes the differential entropy and $\mathbb{E}_{p}[\cdot]$ denotes the expectation over a probability distribution $p$. However, the calculation in Eq. \eqref{ES} is computationally intractable. To resolve this problem, Lobato \emph{et al.} introduced \emph{predictive entropy
search} (PES) by equivalently rewriting Eq. \eqref{ES} as 
\begin{equation}
\operatorname{PES}=\mathrm{H}\left[p\left(y \mid \mathcal{D}_{n}, \mathbf{x}\right)\right]-\mathbb{E}_{p\left(\mathbf{x}^{*} \mid \mathcal{D}_{n}\right)}\left[\mathrm{H}\left[p\left(y \mid \mathcal{D}_{n}, \mathbf{x}, \mathbf{x}^{*}\right)\right]\right].
\end{equation}
Compared with the previous formulation, PES is based on the entropy of predictive distributions, which is analytic or can be easily approximated. Following the same information-theoretic idea, output-entropy-based AFs maximize the reduction of the information about the maximum function value $y^{*}$, the mutual information $I\left(\{\mathbf{x}, y\} ; y^{*} \mid \mathcal{D}_{n}\right)$ instead \cite{wang2017max}. The \emph{max-value entropy search} (MES) is formulated as 
\begin{equation}
\begin{aligned}
&\operatorname{MES}=I\left(\{\mathbf{x}, y\} ; y^{*} \mid \mathcal{D}_{n}\right) \\
&=\mathrm{H}\left(p\left(y \mid \mathcal{D}_{n}, \mathbf{x}\right)\right)-\mathbb{E}_{p\left(y^{*} \mid \mathcal{D}_{n}\right)}\left[\mathrm{H}\left(p\left(y \mid \mathcal{D}_{n}, \mathbf{x}, y^{*}\right)\right)\right].
\end{aligned}
\end{equation}
Intuitively, MES is computationally much simpler than ES and PES as MES uses one-dimensional $p\left(y^{*} \mid \mathcal{D}_{n} \right)$ while ES and PES estimate the expensive and multidimensional $p\left(\mathbf{x}^{*} \mid \mathcal{D}_{n}\right)$. Empirical results have demonstrated that MES performs at least as good as ES and PES.

Note that the above mentioned AFs are all designed for single-objective optimization, and therefore, many recent efforts have been dedicated to developing more new AFs to account for a diverse and wide range of applications. 

\section{Recent Advances in Bayesian Optimization}
\label{Recent Advances}
In the above we provided a brief history of Bayesian optimization, and described the methodology for solving the standard Bayesian optimization problem, i.e., black-box single-objective optimization. 
In this section, we provide an overview of the state-of-the-art Bayesian optimization algorithms, focusing on the most important research advances. In the following, we categorize and discuss the existing work according to the characteristics of the optimization problems to provide a clear picture of the abundant literature.
\subsection{High-dimensional optimization}
High-dimensional black-box optimization problems are extremely challenging yet commonly seen in many applications \cite{wang2018batched,nayebi2019framework}. For example, in hyperparameter optimization of machine learning models \cite{ranjit2019efficient}, the number of hyperparameters and the size of search space grow as the complexity of the models increases. Despite successful applications of Bayesian optimization to low-dimensional expensive and black-box optimization problems, its extension to high-dimensional problems remains a critical open challenge. The dimension of the search space impacts both the construction of GPs and the optimization of AFs. Specifically, the following major difficulties can be identified for Bayesian optimization of high-dimensional problems. 1) Nonparametric regression, such as the GP model, is inherently difficult as the search space grows exponentially with the dimension. On the one hand, it becomes harder to learn a model in a high-dimensional space with the commonly used distance-based kernel functions, as the search spaces grow considerably faster than affordable sampling budgets. On the other hand, the number of hyperparameters generally increases along with the input dimension, as a consequence, the training of the model becomes increasingly hard. 2) Generally, AFs are multi-modal problems, with a large mostly flat surface. Hence, the optimization of AFs is non-trivial, in particular for high-dimensional problems and when the number of samples is limited. For example, Tian \emph{et al.} \cite{tian2019multiobjective} observed that the estimated uncertainty of different solutions in high-dimensional spaces are very similar, reducing the effectiveness of acquisitions based on the estimated uncertainty. Hence, a bi-objective acquisition function was proposed, which is solved using a multi-objective evolutionary algorithm \cite{zhou2011moo}. The reader is referred to Section \ref{sec:MOO} for a more detailed discussion on multi-objective evolutionary algorithms (MOEAs).  

Also, our focus is primarily on the surrogate modeling techniques themselves, not the experimental designs used to generate sample data; interested readers are referred to recent overviews and texts on the topic the GPs are subjected to the scalability challenge
In view of the space limit and the fact that some of strategies are studied in special areas (e.g., parallel computing and increasing computer power), this section only reviews some of them that directly deal with high-dimensionality.
Note that the above problem is related to, but distinct from the scalability of GPs. To construct a reliable GP in higher dimensional space, more observed data may be required, which results in a challenge of scalability for the GP due to its cubic complexity to the data size. Although scalable GPs have been extensively studied in recent years to accommodate many observations \cite{liu2020gaussian}, these methods focus on the scenario where there exist a large amount of data while the dimension remains to be small or medium. Moreover, even if one can fit a GP model for high-dimensional problems, one would still face the difficulty of the optimization of acquisition functions. Therefore, we are interested in scalable Bayesian optimization algorithms for tackling high dimensionality, rather than construction of high-dimensional GPs only.

Most existing Bayesian optimization algorithms for  high-dimensional problems make two structural assumptions, namely low active/effective dimensionality of the objective function together with an additive structure with few exceptions \cite{eriksson2021high}. Addressing high-dimensional Bayesian optimization with a large amount data generally involves alternative models, local modelling, and batch selection in a parallel manner. In the following, we will discuss in detail existing work handling high-dimensional optimization problems.
A comprehensive survey paper in this line can be found in \cite{binois2021survey}, in which the high-dimensional GP modeling is introduced before presenting its application to BO.  

\subsubsection{Variable selection}
To alleviate the curse of dimensionality, a straightforward idea is to adopt a dimension reduction technique. To achieve this, an important assumption often made is that the original objective function varies only within a low-dimensional subspace, called active/effective subspace \cite{chen2012joint}. To identify the most contributing input variables, some sensitivity analysis techniques that evaluate the relative importance of each variable with respect to a quantity of interest have been exploited \cite{spagnol2019global}. In \cite{chen2012joint} two strategies, the finite difference sequential likelihood ratio test and the GP sequential likelihood ratio test, are proposed to screen the most contributing variables. Another commonly used quantity is the values of the correlation lengths of automatic relevance determination (ARD) covariances \cite{winkel2021sequential}. The basic idea is that the larger the length scale value, the less important the corresponding variable. 

\subsubsection{Linear/non-linear embedding}
 Instead of removing the inactive variables to reduce the dimension, more recent developments exploit the active dimensionality of the objective function by defining a latent space based on a linear or non-linear embedding. For example, Wang \emph{et al.} \cite{wang2013bayesian} noted that given any $\mathbf{x} \in \mathbb{R}^{D}$ and a random matrix $\mathbf{A} \in \mathbb{R}^{D \times d}$, at a probability of 1, there is a point $\mathbf{y} \in \mathbb{R}^{d}$ such that $f(\mathbf{x})=f(\mathbf{A y})$. This observation allows us to perform Bayesian optimization in a low-dimensional space to optimize the original high-dimensional function. Hence, an algorithm, called Bayesian optimization with random embedding (REMBO), is proposed. REMBO first randomly generates an embedding denoted as a matrix $A$ and constructs a GP model, and then optimizes an AF to select a new data point that will be projected back to the original space using inverse random projection. Some studies have been dedicated to further investigate the random embedding in recent years. For example, Binois \emph{et al.} \cite{binois2020choice} defined a linear embedding $B$ by inverting the orthogonal projection to relax the bounded region of the embedding in REMBO. Similar to \cite{binois2020choice}, matrix $B$ is used to project the ambient space down to the embedding in \cite{
 letham2020re}. Moreover, the authors gave the reader further insights into the linear embedding for Bayesian optimization in terms of crucial issues and misconceptions, and consequently proposed an adaptive linear embedding Bayesian optimization algorithm. In addition, Nayebi \emph{et al.} \cite{nayebi2019framework} developed an inverse embedding based on a hashing function, and Bayesian optimization with the proposed embedding in combination of a set of AFs has been investigated. Inspired by the studies in \cite{wang2013bayesian}, Cartis \emph{et al.} \cite{cartis2020dimensionality,cartis2021global} extend REMBO to affine subspaces. 
 
Apart from the success in the random embedding methods, many algorithms have been proposed to learn the intrinsic effective subspaces, such as unsupervised learning based on variational auto-encoders \cite{antonova2020bayesian}, supervised learning \cite{zhang2019high}, and semi-supervised learning \cite{siivola2021good}. Specifically, unsupervised dimensionality reduction in the context of high-dimensional Bayesian optimization includes principal component analysis and variational auto-encoders (VAEs). Note that VAEs have emerged as a powerful method for mapping the high-dimensional input space to a low-dimensional latent space. Hence, some research resorts to VAEs to alleviate the scalability issue of Bayesian optimization in high-dimensional input spaces. The early studies for VAE-based Bayesian optimization learn the latent space in an unsupervised way \cite{antonova2020bayesian}. The disadvantage of such a latent space learned using unlabeled data only might be sub-optimal for the optimization task. To address this problem, the most recent work has jointly learned the latent space based on label guidance in a semi-supervised way \cite{siivola2021good}. On the other hand, Zhang \emph{et al.} \cite{zhang2019high} presented a supervised dimension reduction method, sliced inverse regression (SIR), for high-dimensional Bayesian optimization (SIR-BO). SIR-BO performs a heuristic algorithm, i.e., CMA-ES, on the original search space and to optimize the UCB. Alternatively, in \cite{moriconi2020high} two mappings, a non-linear feature mapping to reduce the dimensionality of inputs and a reconstruction mapping to evaluate the objective function, are learned in a supervised learning manner. Consequently, the AF can be optimized in the lower-dimensional feature space. In \cite{chen2020semi}, an extension of SIR using semi-supervised discriminant analysis is proposed, called semi-SIR, to incorporate both labeled and unlabeled points acquired from the acquisition function of Bayesian optimization to learn the embedding space. 

Most of the above-mentioned methods based on the structural assumption use linear projections to scale Bayesian optimization to high dimensions. Recently, a few advanced techniques have been developed to further investigate the structure of the search space by using non-linear embeddings. Compared with linear embeddings, non-linear embedding techniques, also known as geometry-aware Bayesian optimization \cite{oh2018bock}, can be considerably more expressive and flexible. However, these methods require even more data to learn the embedding and assume that the search space is not Euclidean but various manifolds, such as Riemannian manifold \cite{jaquier2019bayesian}. The Euclidean geometry of the search space incurs the boundary issue where an algorithm over-explores the boundaries of the search space, especially in high dimensions. Under this observation, Oh \emph{et al.} \cite{oh2018bock} applied a cylindrical geometric transformation on the search space, resulting in a new kernel, referred to as the cylindrical kernel. By leveraging the new kernel in Bayesian optimization, called BOCK, one can easily scale it to more than 50 dimensions and mitigate the boundary issue. Arguably, BOCK is the first work where geometry-awareness is considered within the framework of Bayesian optimization. 

Another seminal avenue is built on Riemannian manifold theory. For applications with non-Euclidean search spaces, such as robotics \cite{jaquier2019bayesian}, the Euclidean methods can be quite brittle, motivating the recently developed geometry-aware Bayesian optimization. Geometry-awareness is introduced into Bayesian optimization to exploit the manifold's geometric structure, so that the GP can properly measure the similarities of the non-Euclidean parameter space, with the hope of improving Bayesian optimization's performance and scalability. To achieve this, Jaquier \emph{et al.} \cite{jaquier2019bayesian} proposed a geometry-aware Bayesian optimization (GaBO) with new kernels measuring the similarities of manifold-valued data, and two commonly used manifolds in robotics have been considered, i.e., the sphere and symmetric positive definite manifolds. Moreover, the optimization of AFs is performed on Riemannian manifolds. A subsequent work by Jaquier and Rozo \cite{jaquier2020high} extends GaBO to high-dimensional problems, namely HD-GaBO, by learning a nested structure-preserving mapping from the original manifold to a lower-dimensional latent space. In HD-GaBO, the mapping and the objective function in the latent space are jointly learned using a manifold Gaussian process (mGP) with the geometry-aware kernel function. It is necessary to investigate mathematical theory and techniques for building new kernels in the manifold settings, since the naive geometric approximations may lead to ill-defined kernels. To address this problem, Borovitskiy \emph{et al.} \cite{borovitskiy2020mat} provided mathematically sound techniques for computing the geometry-aware kernels in the Riemannian setting via Laplace–Beltrami eigenfunctions. The most recent work by Jaquier \emph{et al.} \cite{jaquier2021geometry} has extend the GaBO with the theoretically-grounded Matérn kernels proposed in \cite{borovitskiy2020mat} in robotics. 
 
\subsubsection{Addictive structure}
The low active dimensionality assumption behind the aforementioned methods is too restrictive as all the input variables may contribute to the objective function. Hence, another salient structure assumption, called addictive structure, has been explored in the context of high-dimensional Bayesian optimization. Note that the addictive structure has been widely used in addictive models \cite{lin2020transformation,chen2020projection} and kernel functions. 
A prominent work in the context of Bayesian optimization and bandits, namely 
Add-GP-UCB, was proposed in \cite{kandasamy2015high}, assuming that the objective function is a sum of functions of small, disjoint groups of dimensions. Instead of directly using addictive kernels, a set of latent decompositions of the feature space is generated randomly and the one with the highest GP marginal likelihood is chosen, with each kernel operating on subsets of the input dimensions. Markov Chain Monte Carlo (MCMC) \cite{gardner2017discovering}, Gibbs sampling \cite{wang2017batched} and Thompson sampling \cite{wang2017max} were also introduced to more effectively learn the addictive structure. A recent work by Delbridge \emph{et al.} \cite{delbridge2020randomly} presents a learning-free addictive GP based on sequences of multiple random projections to avoid the computationally expensive learning for the addictive structure.

Another major issue concerning Add-GP-UCB is the restriction of disjoint subsets of input dimensions, which have been lifted in subsequent work \cite{li2016high,rolland2018high}. Li \emph{et al.} generalized the two structure assumptions, i.e., the low active assumption and the addictive structure assumption, by introducing a projected-addictive assumption. In \cite{rolland2018high}, overlapping groups are allowed by representing the addictive decomposition via a dependency graph or a sparse factor graph.  

\subsubsection{Large-scale data in high-dimensional Bayesian Optimization}
While there have been ample studies on Bayesian optimization to account for problems with large-scale observations and high-dimensional input spaces, very few have considered high-dimensional problems with a large amount of training data. This optimization scenario is indispensable as more data is required for constructing surrogates in high-dimensional spaces. Earlier research has shed some light on the potential advantages of replacing GPs with more scalable and flexible machine learning models. A natural choice is Bayesian neural networks due to their desirable flexibility and characterization of uncertainty \cite{springenberg2016bayesian}. Guo \emph{et al.} \cite{guo2021evolutionary} developed an efficient dropout neural network (EDN) to replace GPs in high-dimensional multi/many-objective optimization. The core idea in EDN is that the dropout is executed during both the training and prediction processes, so that EDN is able to estimate the uncertainty for its prediction. Alternatively, random forests have been adopted to replace GPs to address large-scale high-dimensional problems, as done in \cite{hutter2011sequential}. More recently, a few methods have proposed that resort to local modelling and batch selection in a parallel manner to scale Bayesian optimization to problems with large-scale observations and high-dimensional input spaces. Wang \emph{et al.} \cite{wang2018batched} proposed ensemble Bayesian optimization (EBO) to alleviate the difficulties of constructing GPs and optimizing AFs for high-dimensional problems. EBO firstly learns local models on partitions of the input space and subsequently leverages the batch selection of new queries in each partition. Similarly,  an MOEA with a heterogeneous ensemble model as a surrogate was proposed \cite{guo2018heterogeneous}, in which each member is trained by different input features generated by feature selection or feature extraction. The trust region method is adopted to design a local probabilistic approach (namely TuRBO) for handling large-scale data in high-dimensional spaces \cite{eriksson2019scalable}. However, the trust regions in TuRBO are learned independently without sharing data, which may be inefficient for expensive problems. To address this issue, a data-sharing strategy is introduced and TuRBO is extended to MOPs by employing an AF based on hypervolume improvement \cite{daulton2021multi}.

\subsection{Combinatorial optimization}
The optimization of black-box functions over combinatorial spaces, e.g., integer, sets, categorical, or graph structured input variables, is ubiquitous and yet challenging task in real-world science and engineering applications. Without loss of generality, suppose there is an expensive black-box objective function $f:\mathcal{H} \rightarrow \mathbb{R}$. The goal of combinatorial optimization is:
\begin{equation}
\mathbf{h}^{*}=\arg \max f(\mathbf{h})
\end{equation} 
where $\mathcal{H}$ denotes the search space. For problems over a hybrid search space, $\mathcal{H}=[\mathcal{C}, \mathcal{X}]$, $\mathcal{C}$ and $\mathcal{X}$ denote the categorical and continuous search space, respectively. For problems over categorical domains, we simply have $\mathcal{H}=\mathcal{C}$. 

Bayesian optimization has emerged as a well-established paradigm for handling costly-to-evaluate black-box problems. However, most Gaussian process-based Bayesian optimization algorithms explicitly assume a continuous space, incurring poor scalability to combinatorial domains. Moreover, Bayesian optimization suffers seriously from the fact that the number of possible solutions grows exponentially with the parameters in the combinatorial domain (known as combinatorial explosion). Consequently, there are two major challenges for combinatorial Bayesian optimization. One is the construction of effective surrogate models over the combinatorial space, and the other is the effective search in the combinatorial domain for the next structure for evaluation according to the acquisition function. A straightforward way is to construct GPs and optimize AFs by treating discrete variables as continuous, and then the closest integer for the identified next sample point with real values is obtained via a one-hot encoding strategy \cite{garrido2020dealing}. Clearly, this approach ignores the nature of the search space and may repeatedly select the same new samples, which deteriorates the efficiency of Bayesian optimization. Alternatively, many studies borrowed the elegance of VAEs to map high-dimensional, discrete inputs onto a lower dimensional continuous space \cite{gomez2018automatic}. In the context of Bayesian optimization, much effort has been dedicated to handling expensive combinatorial optimization problems by introducing surrogate models for combinatorial spaces. 

\subsubsection{Latent representation}
Instead of carefully measure the similarity in the discrete space, exploring the continues latent space provides an appealing approach to develop the combinatorial BO with GP models. Many studies borrowed the elegance of variational autoencoders (VAEs), a encoder-decoder style deep generative model from machine learning community, to map high-dimensional, discrete inputs to a lower dimensional continuous space. The VAE with a SMILES encoder has been first adopted in \cite{gomez2018automatic} by Bombarelli \emph{et al.} to handle the optimization in a discrete and large molecular space. However, the decoder generate invalid moleculars due to the internal syntax of the SMILES encoder, which is called the decoding error issue. Hence, some follow-up works \cite{griffiths2020constrained} has been developed recently to tackle this issue. Motivated by the fact that surrogate models over the latent space only utilize the information learned by the VAE, Deshwal and Doppa \cite{deshwal2021combining} suggested to leverage both the structural information in the original space and the latent space representations, which is achieved by a structure-coupled kernel. Besides, Reproducing Kernel Hilbert Space embedding \cite{buathong2020kernels} and random embedding \cite{kim2020combinatorial} also have been used to construct the latent space in the context of combinatorial BO. 
\subsubsection{One-hot transformation}
Many efforts have been dedicated \seb{to} handling expensive combinatorial optimization problems in the context of BO. A straightforward way is to construct GPs and optimise AFs by treating discrete variables as continuous, and then the closest integer for the identified next sample point with real values is obtained via a one-hot encoding strategy \cite{garrido2020dealing}. Clearly, this approach ignores the nature of the search space and may repeatedly select the same new samples, which deteriorates the efficiency of BO. 

\subsubsection{Inherently discrete models} 
To sidestep the difficulties encountered in the GP-based Bayesian optimization, some inherently discrete models (e.g. neural networks \cite{swersky2020amortized} and random forests) are employed as surrogate models, among which tree-based models are the most widely used ones. For example, random forests have been applied to the combinatorial Bayesian optimization in \cite{hutter2010sequential}. Unfortunately, this approach suffers from performing undesirable extrapolation. Hence, a tree-structured Parzen estimator (TPE) model has been used to replace the GPs in \cite{bergstra2013hyperopt}, which, however, requires a large number of training data. An alternative idea is to use continuous surrogate models that guarantee integer-valued optima, which motivates a method called IDONE \cite{bliek2021black} using a piece-wise linear surrogate model. 

To improve the search efficiency of the AF in combinatorial optimization, search control knowledge is introduced to branch-and-bound search \cite{deshwal2020optimizing}. In addition, an algorithm called BOCS is proposed to alleviate the combinatorial explosion of the combinatorial space \cite{baptista2018bayesian}. 
In BOCS, a sparse Bayesian linear model is used to handle the discrete structured domain, and the selection of new sample points is formulated as a semi-definite program. However, BOCS can be prohibitive for a large number of binary and categorical variables due to the one-hot encoding representation. To address this issue, a subsequent work is developed using a submodular relaxation \cite{deshwal2020scalable}. 

a tree-structured Parzen estimator (TPE) model, an Estimation of Distribution based approach, has been used to replace the GPs in \emph{et al.} \cite{bergstra2011algorithms,bergstra2013hyperopt}

\subsubsection{Kernels with discrete distance measures}
Another popular avenue for combinatorial Bayesian optimization is to modify the distance measure in the kernel calculation, so that the similarity in the combinatorial space can be properly captured. For example, the Hamming distance is widely used to measure the similarity between discrete variables, and an evolutionary algorithm is generally adopted to optimize the AF \cite{hutter2010sequential}. More recently, graph presentations of combinatorial spaces has emerged at the forefront, contributing to graph kernels in GPs. Oh \emph{et al.} \cite{oh2019combinatorial} proposed COMBO, which constructs a combinatorial graph over the combinatorial search space, in which the shortest path between two vertices in the graph is equivalent to the Hamming distance. Subsequently, graph Fourier transforms are utilized to derive the diffusion kernel on the graph. To circumvent the computational bottleneck of COMBO,  the structure of the graph representation is further studied and a small set of features is extracted \cite{deshwal2020mercer}. Note that graph-based combinatorial Bayesian optimization has been widely applied to neural architecture search \cite{kandasamy2018neural,ru2021interpretable}. 

\subsubsection{Bayesian optimization over mixed search spaces}
Very few studies have considered mixed-variable combinatorial problems, where the input variables involve both continuous and discrete ones, such as integers and categorical inputs. The kernels with new distance measures over discrete spaces have shed light on addressing combinatorial optimization problems. Hence, some attempts have been made for combinatorial Bayesian optimization in a similar fashion, i.e., combining kernels defined over different input variables \cite{ru2020bayesian}. While replacing the GPs in the framework of Bayesian optimization is a possible approach in the mixed-variable setting \cite{bliek2021black}, the bandit approaches have been integrated with Bayesian optimization by treating each variable as a bandit \cite{nguyen2020bayesian}.

\subsection{Noisy and robust optimization}
Two assumptions about the noise in the data are made for constructing the GP in Bayesian optimization \cite{mchutchon2011gaussian}. First, the measurement of the input points is noise-free. Second, noise in observations is often assumed to follow a constant-variance normal distribution, called homoscedastic Gaussian white noise. However, neither of these assumptions may hold in practice, rendering poor optimization performance. Hence, Bayesian optimization approaches accounting for noisy observations, outliers, and input-dependent noise have been developed.

\subsubsection{Bayesian optimization for output noise} 
For an optimization with noisy output, the objective function can be described by $f: \mathcal{X} \rightarrow \mathbb{R}$ resulting from noisy observations $y=f\left(\mathbf{x}\right)+\epsilon$, where $\epsilon$ is addictive/output noise.
For example, some stochastic simulations for objective evaluations involve finite element analysis \cite{zhang2020bayesian}, density functional theory \cite{vargas2020bayesian}, Monte Carlo simulation \cite{picheny2013quantile} and discrete event simulation \cite{zmijewski2020evaluation}, which, if repeated, will give different results. 
Most Bayesian optimization approaches for problems in the presence of output noise employ the standard GP as the surrogate model and focus on designing new AFs \cite{picheny2013benchmark}. Firstly, the extension of the noise-free EI (Eq. \ref{EI}) to noisy observations has been studied extensively \cite{zhan2020expected}. One major issue is that the current best objective value $f(\mathbf{x^*})$ is not exactly known. A direct approach is to replace $f(\mathbf{x^*})$ by some sensible values, which is called expected improvement with “plug-in" \cite{picheny2013benchmark}. For example, Vazquez \emph{et al.} \cite{vazquez2008global} used the minimum of the GP prediction as $f(\mathbf{x^*})$ to derive a modified EI. However, it does not degenerate to the standard EI when the output noise goes to zero. Hence, Huang \emph{et al.} \cite{huang2006global} developed an augmented EI by replacing the current best objective value and subsequently added a penalty term to the standard EI. Alternatively, the $\beta$-quantile given by the GP surrogate is used as a reference in \cite{picheny2013quantile}. In that work, an improvement based on the decrease of the lowest of the $\beta$-quantile is further defined, yielding the expected quantile improvement (QEI) that is able to account for heterogeneous noise. Similar to QEI, the improvement is defined by the knowledge gradient (KG) policy, and an approximate knowledge gradient (AKG) is introduced \cite{scott2011correlated}. Fundamentally, AKG is an EI based on the knowledge improvement; however, the evaluation of AKG is computationally intensive. A thorough introduction and comparison of different noisy AFs, especially variants of EI, can be found in \cite{picheny2013benchmark, jalali2017comparison}. 
Another class of AFs that naturally handles output noise is information-based AFs, such as the predictive entropy search \cite{hernandez2014predictive} and Thompson sampling algorithm \cite{kandasamy2018parallelised}.

A reinterpolation method was also proposed to handle output noise \cite{forrester2006design}, where a Kriging regression is constructed using noisy observations. Then, the sampled points with the predictions provided by the Kriging are adopted to build an interpolating Kriging, which is called the reinterpolation, enabling the standard EI to select new samples. The reinterpolation has been extend to multi-objective optimizations in \cite{koch2015efficient}. 

\subsubsection{Bayesian Optimization for outliers}
Besides the above mentioned measurement/output noise, the observations are often contaminated with outliers/extreme observations in real experiments due to irregular and isolated disturbances, instrument failures, or potential human errors. Take multi-class classification problems as an example, some training data points may be misplaced on the wrong side in the feature space. The hyperparameter tuning may also encounter outliers due to the code bug or a network issue. 
As pointed out in O'Hagan \cite{o1979outlier}, the standard GP model that adopts Gaussian distributions as both the prior and the likelihood is sensitive to extreme observations. 

To account for the outliers in the observations, robust GP models that are insensitive to the presence of outliers have been developed. Mathematically, the main idea behind robust GP models is to use an appropriate noise model with a heavier tail, instead of assuming normal noise, to account for the outlying data \cite{kodamana2018approaches}. A straightforward model is the weighted convex combination of a regular Gaussian distribution with a relatively small variance for regular observations and a wide Gaussian distribution with a large variance for extreme observations \cite{kodamana2018approaches}. The Gaussian distribution with a moderate variance indicates that an observation is considered as a regular measurement with a high probability, while the wide Gaussian distribution with a larger variance assumes that the occurrence of extreme outliers cannot be denied. Note that the posterior of the mixture likelihood cannot be computed analytically as it is no longer a Gaussian distribution.
In \cite{kuss2005approximate}, the expectation-propagation approximation and Markov chain Monte Carlo techniques are adopted for the approximate posterior inferences.  
The most commonly used noise model is Student-t distribution \cite{vanhatalo2009gaussian,martinez2018practical}. The probability density function of the Student-t distribution is formulated as:
\begin{equation}
p\left(\epsilon \mid \mu, \sigma^{2}, v\right)=\frac{\Gamma((\nu+1) / 2)}{\Gamma((\nu / 2)) \sqrt{v \pi \sigma^{2}}}\left(1+\frac{1}{v}\left(\frac{\epsilon-\mu}{\sigma}\right)^{2}\right)^{-\frac{\nu+1}{2}}
\end{equation}
where $\mu$ is the mean of the distribution, $\sigma>0$ represents a scaling parameter, and $\nu>0$ denotes the degree of freedom to control the thickness of the tails in the distribution. It is clear that using the Student-t likelihood will not allow a closed form of inference of the posterior distribution, therefore, some techniques of approximate inference are required. For example, Kuss \cite{Kuss:06} adopted a factorizing variational approximation (fVB) and an alternative Markov Chain Monte Carlo scheme to implement approximate inference in the GP model with a Student-t likelihood. Another attempt for approximate inference is the Laplace approximation \cite{vanhatalo2009gaussian}. Motivated by the prior work \cite{minka2001family}, expectation propagation (EP), a method to approximate integrals over functions that factor into simple terms, is adopted to handle the approximate inference problem rendered by the Student-t likelihood \cite{martinez2017robust}. 
More recently, Martinez-Cantin \cite{martinez2018practical} proposed an outlier-handling algorithm by combining a robust GP with Student-t likelihood with outlier diagnostics to classify data points as outliers or inliers. Thus, the outliers can be removed and a standard GP can be performed, resulting in a more efficient robust method with a better convergence.

The GP model combined with the Student-t noise model makes the inference problem challenging due to the potential multimodality of the posterior caused by the non-log-concave Student-t likelihood. As stated in \cite{kuss2006gaussian}, the likelihood has to be log-concave to guarantee its modelling unimodality of the posterior in GPs. Hence, Laplace noise is a notable choice for the likelihood owing to its sharp peaks and longer and fatter tails  while still being log-concave. Alternatively, Flat-topped t-distribution has been investigated to take the uncertainty of the noise into consideration \cite{kodamana2018approaches}.

\subsubsection{Bayesian optimization for corrupted inputs}
As we mentioned before, Bayesian optimization is intrinsically robust to noisy function evaluations, because the standard GP typically assumes that the output measurement is corrupted by noise, regardless of the input vector. 
The input-dependent noise was first considered in modeling GP \cite{goldberg1997regression}, where heteroscedastic noise was introduced by allowing the noise variance to be a function of input instead of a constant. Hence, the noise variance is considered as a random variable and an independent GP is used to model the logarithms of the noise level. The inference in heteroscedastic GP (HGP) regression is challenging, since, unlike in the homoscedastic case, the predictive density and marginal likelihood are no longer analytically tractable. The MCMC method can be used to approximate the posterior noise variance, which is, however, is time-consuming. Suggested alternative approximations include variational inference \cite{lazaro2011variational}, Laplace approximation and expectation propagation. Kersting \emph{et al.} \cite{kersting2007most} developed a maximum-a-posteriori approach. The author pointed out that  the algorithm is not guaranteed to converge and may instead oscillate as it considers most-likely completions of the data only.Similar studies include \cite{le2005heteroscedastic,quadrianto2009kernel,yuan2004doubly}.

The above mentioned methods handle datasets with input noise by holding the input measurements as deterministic and changing the corresponding output variance to compensate. McHutchon and Rasmussen \cite{mchutchon2011gaussian} pointed out that the effect of the input-dependent noise is related to the gradient of the function mapping input to output. Therefore, a noisy input GP (NIGP) was developed, where the input noise is transferred to output based on a first order Taylor expansion of the posterior. Specifically, NIGP adopts a local linearization of the function, and uses it to propagate uncertainty from the inputs to the output of the GP \cite{mchutchon2011gaussian}. The NIGP is particularly effective for time-series data where the output at time $t-1$ becomes the input at time $t$. However, it is designed to tackle constant-variance input noise. 
 
The intuition behind the above ideas is to propagate the input noise to the output space, which may, however, result in unnecessary exploration. Nogueira \emph{et al.} \cite{nogueira2016unscented} addressed this issue by considering input noise in EI, so that the input noise can be propagated through all the models and the function queries. More precisely, an unscented expected improvement and an unscented optimal incumbent are defined using the unscented transformation (UT). UT first deterministically chooses a set of samples from the original distribution. Then, a nonlinear function is applied to each sample to yield transformed points. Hence, the mean and covariance of the transformed distribution can be formed according to the weighted combination of the transformed points. 

A closely related term to input-dependent noise is input/query uncertainty \cite{beland2017bayesian}. That is, the estimation of the actual query location is also subject to uncertainty, such as environmental variables \cite{marzat2013worst} or noise-corrupted inputs. When extending Bayesian optimization to problems with input uncertainty, two classical problem formulations, a probabilistic robust optimization and worst-case robust optimization, from a probabilistic and deterministic point of view have been adopted \cite{kirschner2020distributionally}. In probabilistic robust optimization, a distribution of the input or environmental variables is assumed. Hence, a prior is placed on the input space in order to account for localization noise, and performance is assessed by the expected value of some robustness measurement. A representative work by Bland and Nair \cite{beland2017bayesian} introduces noise-corrupted inputs, namely uncertainty, within the framework of Bayesian optimization. In this case, a robust optimization problem is formulated as a constrained problem by integrating an unknown function with respect to the input distributions. Hence, the noise factors can be integrated out and an AF similar to the constrained EI is introduced to select new queries entirely in the decision space. More recently, such a robust Bayesian optimization setting has been studied by Fröhlich \emph{et al.} \cite{frohlich2020noisy}, where a noisy-input entropy search (NES) based on MES is proposed. BO under input uncertainty, i.e., the estimation of the actual query location is also subject to uncertainty, has been applied to the optimization of stochastic simulations \cite{wang2020gaussian} and the action choice \cite{kirschner2019stochastic}.  

By contrast, the worst-case robust objective aims to search for a solution that is robust to the worst possible realization of the uncertain parameter, which is formulated as a min-max optimization problem,
\begin{equation}
\max _{\mathbf{x}} \min _{\mathbf{c} \in U} f(\mathbf{x}, \mathbf{c}),
\end{equation}
where $\mathbf{x}$ denotes the decision vector, $\mathbf{c} \in U$ denotes uncertainties, where $U$ is the uncertainty set. Marzat \cite{marzat2013worst} uses a relaxation procedure to explore the use of EGO for worst-case robust optimization, so that the design variables and the uncertainty variables can be optimized iteratively. However, such a strategy is inefficient as the previous observations are not reused. Ur Rehman \emph{et al.} \cite{ur2014efficient} proposed a modified EI using a new expected improvement. Some more complex problem settings have been studied in worst-case context, including  distributionally robust Bayesian optimization \cite{kirschner2020distributionally} and adversarial corruptions \cite{bogunovic2020corruption}.

If, as an approximation, we treat the input measurements as if they were deterministic, and inflate the corresponding output variance to compensate, this leads to the output noise variance varying across the input space, a feature often called heteroscedasticity. Unfortunately, in the GP framework, considering each input location to be a distribution is intractable. Noisy-input entropy search for efficient robust bayesian optimization
\subsection{Expensive constrained optimization}
Many optimization problems are subject to various types of constraints, and the evaluation of both the objective function and the constraints can be computationally intensive or financially expensive, known as expensive constrained optimization problems (ECOPs). For example, in control systems the tuning of PID controllers aims to optimize the performance indicator while guaranteeing the stability and safety \cite{konig2020safety}. 
Without loss of generality, an ECOP can be formulated as 
\begin{equation}
\begin{array}{ll}
\min _{\mathbf{x}} & \mathbf{f}(\mathbf{x})=\left(f_{1}(\mathbf{x}), \ldots, f_{m}(\mathbf{x})\right) \\
\text  { s.t. } & c_{j}(\mathbf{x}) \geq a_{j}, \quad j=1, \ldots, q  \\
& \mathbf{x} \in X
\end{array}
\end{equation}
where $\mathrm{x}=\left(x_{1}, x_{2}, \ldots, x_{d}\right)$ is the decision vector with $d$ decision variables, $X$ denotes the decision space, $c_{j}(\mathbf{x})$ is the $j$-th inequality and equality constraints, respectively. Since we consider both single-objective and multi-objective problems, the objective vector $f$ consists of $m$ objectives and $m=1, 2, \cdots, N$. In this setting, only solutions contained in the feasible space defined by the constraints are valid, called feasible solutions. Consequently, the optimization becomes more challenging in the presence of constraints.

Indeed, the past decades have seen the rapid development of constraint-handling techniques in many fields, especially in the evolutionary computation community. However, most methods are untenable in the presence of expensive objectives and constraints, which motivates a proliferation of studies exploring the use of Bayesian optimization for ECOPs. A natural idea to account for constraints is to use the augmented Lagrangian relaxation to convert constrained problems into simple unconstrained problems and then Bayesian optimization can be applied directly \cite{gramacy2016modeling}. 
Bayesian optimization for constrained optimization problems can be roughly classified into two groups. 1) With the help of GPs, new acquisition functions are proposed to account for the constraints within the framework of Bayesian optimization, known as constrained Bayesian optimization (CBO). Recently, CBO has become popular, especially for addressing single-objective constrained problems. According to the different acquisition functions in CBO, we classify various CBO algorithms into three sub-categories: probability of feasibility based, expected volume reduction based, and multi-step look-ahead methods. 2) To circumvent the computational burden encountered in ECOPs, Bayesian optimization is adopted in existing constraint-handling methods, typically, evolutionary algorithms. We refer to these as surrogate-assisted constraint-handling methods. In the following, each group is introduced and discussed.

\textbf{Augmented Lagrangian relaxation}:
A natural and straightforward idea to account for constraints is to convert constrained problems into simple unconstrained problems. This can be achieved by the augmented Lagrangian (AL), given by
\begin{equation}
L_{A}(\mathbf{x}; \lambda, \rho)=f(\mathbf{x})+\lambda^{\top} c(\mathbf{x})+\frac{1}{2 \rho} \sum_{j=1}^{q} \max \left(0, c_{j}(\mathbf{x})\right)^{2}
\end{equation}
where $\rho>0$ is a penalty parameter and $\lambda \in \mathbb{R}_{+}^{q}$ denotes Lagrange multiplier. Intuitively, a surrogate model can be adopted to directly model the AL. However, as pointed out in \cite{gramacy2016modeling}, this way requires nonstationary surrogate models, thereby resulting in modeling difficulties. Instead, the authors separately modeled the objectives and constraints, and constructed an inner AL subproblem. Hence, the EI can be applied by replacing the current best observation with the current best value of the AL. This work has been extended to expensive problems with mixed constraints in \cite{picheny2016bayesian}, where an alternative slack variable AL is proposed by introducing slack variables. More recently, an Alternating Direction Method of Multipliers (ADMM) based on the AL function has been cooperated with BO to effectively handle problems subject to unknown constraints in \cite{ariafar2019admmbo}.

\subsubsection{Probability of feasibility}
The combination of the existing AFs with feasibility indicators, such as probability of feasibility, offers a principled approach to constrained optimization. The most representative work is the extension of the well-established EI, known as EI with constraints (EIC) \cite{griffiths2020constrained,tran2020srmo}. One of the previous EIC methods, called constrained EI (cEI) or constraint-weighted EI, aims to maximize the expected feasible improvement over the current best feasible observation. Typically, cEI multiplies the EI and the constrained satisfaction probabilities, formulated as follows:
\begin{equation}
\mathrm{cEI}(\mathbf{x})=EI(\mathbf{x})\prod_{j=1}^{q} \operatorname{Pr}\left(c_{j}(\mathbf{x}) \leq a_{j}\right)
\label{cEI}
\end{equation}
where each constraint is assumed to be independent, and all expensive-to-evaluate functions are approximated by independent GPs. Interestingly, similar ideas have been discussed in \cite{schonlau1998global} and revisited in \cite{gardner2014bayesian}. As indicated in Equation (\ref{cEI}), cEI faces several issues. First, the current best observation is required, which is untenable in some applications, such as noisy experiments. Hence, a recent work by Letham \emph{et al.}  \cite{letham2019constrained} directly extends cEI to noisy observations with greedy batch optimization. Second, cEI can be brittle for highly constrained problems \cite{letham2019constrained}.  

As a promising selection criterion in the presence of constraints, EIC has been studied in a variety of settings \cite{griffiths2020constrained,konig2020safety}. 
Note that EIC has been extended to multi-objective optimization by introducing the Pareto dominant probability \cite{tran2020srmo}. The unknown constraints have been taken into consideration in \cite{tran2020srmo}. More recently, a new variant of the knowledge gradient has been proposed to account for constraints using the probability of feasibility \cite{ungredda2021bayesian,chen2021new}. 

\subsubsection{Expected volume reduction} 
Another class of AFs is derived to accommodate constraints by reducing a specific type of uncertainty measure about a quantity of interest based on the observations, which is known as stepwise uncertainty reduction \cite{chevalier2014fast}. As suggested in previous studies \cite{chevalier2014fast}, many AFs can be derived to infer any quantity of interest, depending on different types of uncertainty measures. In \cite{picheny2014stepwise}, an uncertainty measure based on PI has been defined, where constraints are further accounted for by combining the probability of feasibility. The most recent work \cite{amri2021sampling} has revisited this idea, and a new uncertainty measure is given by the variance of the feasible improvement. Using the same principle, integrated expected conditional improvement (IECI) in \cite{bernardo2011optimization} defines the expected reduction in EI under the constrained satisfaction probabilities, allowing the unfeasible area to provide information. Another popular uncertainty measure is entropy inspired by information theory, which has been explored in \cite{hernandez2014predictive,perrone2019constrained}. Hern{\'a}ndez-Lobato \emph{et al.} \cite{hernandez2015predictive} extended Predictive Entropy Search (PES) to unknown constrained problems by introducing the conditional predictive distributions, with the assumption of the independent GP priors of the objective and constraints. A follow-up work \cite{hernandez2016general} further investigated the use of PES in the presence of decoupled constraints, in which subsets of the
objective and constraint functions can be evaluated independently. However, PES encounters the difficulty of calculation, which motivates the use of max-value entropy search for constrained problems in a recent work \cite{perrone2019constrained}.

\subsubsection{Multi-step look-ahead methods} 
Most AFs are myopic, called one-step look-ahead methods, as they greedily select locations for the next true evaluation, ignoring the impact of the current selection on the future steps. By contrast, few non-myopic AFs have been developed to select samples by maximizing the long-term reward from a multi-step look-ahead \cite{yue2020non}. For example, Lam and Willcox \cite{lam2016bayesian} formulated the look-ahead Bayesian optimization as a dynamic programming (DP) problem, which is solved by an approximate DP approach called rollout. This work subsequently was extended to constrained Bayesian optimization by redefining the stage-reward as the reduction of the objective function satisfying the constraints \cite{lam2017lookahead}. The computation burden resulting from rollout triggers the most recent work by Zhang \emph{et al.} \cite{zhang2021constrained}, where a constrained two-step AF, called 2-OPT-C, has been proposed. Moreover, the likelihood ratios method is used to effectively optimize 2-OPT-C.  
 It is worth noting that, this can be partially achieved by batch BO algorithms capable of jointly optimizing a batch of inputs because their selection of each input has to account for that of all other inputs of the batch. However, since the batch size is typically set to be much smaller than the given budget, they have to repeatedly select the next batch greedily. 
As a promising selection paradigm, the multi-step lookaheand BO has been explored to account for constraints recently.
\subsubsection{Surrogate-assisted constraint-handling methods}
The above-mentioned constraint-handling techniques focus on the AFs within the Bayesian optimization framework, where a GP model generally serves as a global model. In the evolutionary computation community, many attempts have been made to combine the best of both worlds in the presence of expensive problems subject to constraints. One avenue is to use MOEAs to optimize the objectives and constraints simultaneously. For example, instead of maximizing the product of EI and the probability of feasibility, the two AFs can be served as two objectives and optimized by an MOEA \cite{zhang2021efficient}. In case there is no feasible solutions, another common method proposed to search the feasible region first, and then approaches to the best feasible solution. Moreover, it is difficult to construct surrogates with good quality using limited training data. Hence, conducting both local and global search has attracted much attention recently \cite{jiao2019complete,akbari2020kasra,cheng2021parallel}. 


\subsection{Multi-objective optimization}
\label{sec:MOO}
Many real-world  optimization  problems  have  multiple conflicting objectives to be optimized simultaneously, which are referred to as multi-objective optimization problems (MOPs) \cite{zhou2011moo}. Mathematically, an MOP can be formulated as 
\begin{equation}
\begin{array}{ll}
\min _{\mathbf{x}} & \mathbf{f}(\mathbf{x})=\left(f_{1}(\mathbf{x}), f_{2}(\mathbf{x}), \ldots, f_{m}(\mathbf{x})\right) \\
\text { s.t. } & \mathbf{x} \in \mathcal{X}
\end{array}
\end{equation}
where $\mathbf{x}=\left(x_{1}, x_{2}, \ldots, x_{d}\right)$ is the decision vector with $d$ decision variables, $\mathcal{X}$ denotes the decision space, and the objective vector $\mathbf{f}$ consists of $m$ $(m\geq2)$ objectives. Note that for many-objective problems (MaOPs) \cite{li2015maop}, the number of objectives $m$ is larger than three. Here the target is to find a set of optimal solutions that trade off between different objectives, which are known as Pareto optimal solutions. The whole set of Pareto optimal solutions in the decision space is called Pareto set (PS), and the projection of PS in the objective space is called Pareto front (PF). The aim of multi-objective optimization is to find a representative subset of the Pareto front and MOEAs have been shown to be successful to tackle MOPs \cite{zhou2011moo}. 


Like single-objective optimization, the objective functions in an MOP can be either time-consuming or costly. Some examples include airfoil design, manufacturing engineering, the design of crude oil distillation units, and furnace optimization. 
Thus, only a small number of fitness evaluations is affordable, making plain MOEAs hardly practical. Recall that GPs and AFs in Bayesian optimization are designed for single-objective black-box problems, therefore new challenges arise when Bayesian optimization is extended to MOPs, where sampling of multiple objective functions needs to be determined, and both accuracy and diversity of the obtained solution set must be taken into account. To meet these challenges, multi-objective Bayesian optimization is proposed by either embedding Bayesian optimization into MOEAs or converting an MOP into single-objective problems. Multi-objective Bayesian optimization can be largely divided into three categories: combinations of Bayesian optimization with MOEAs, performance indicator based AFs, and information theory based AFs. Note that some of them may overlap and are thus not completely separable.

\subsubsection{Combinations of Bayesian optimization with MOEAs} 
\begin{figure}[ht]
\centering
\includegraphics[width=\columnwidth]{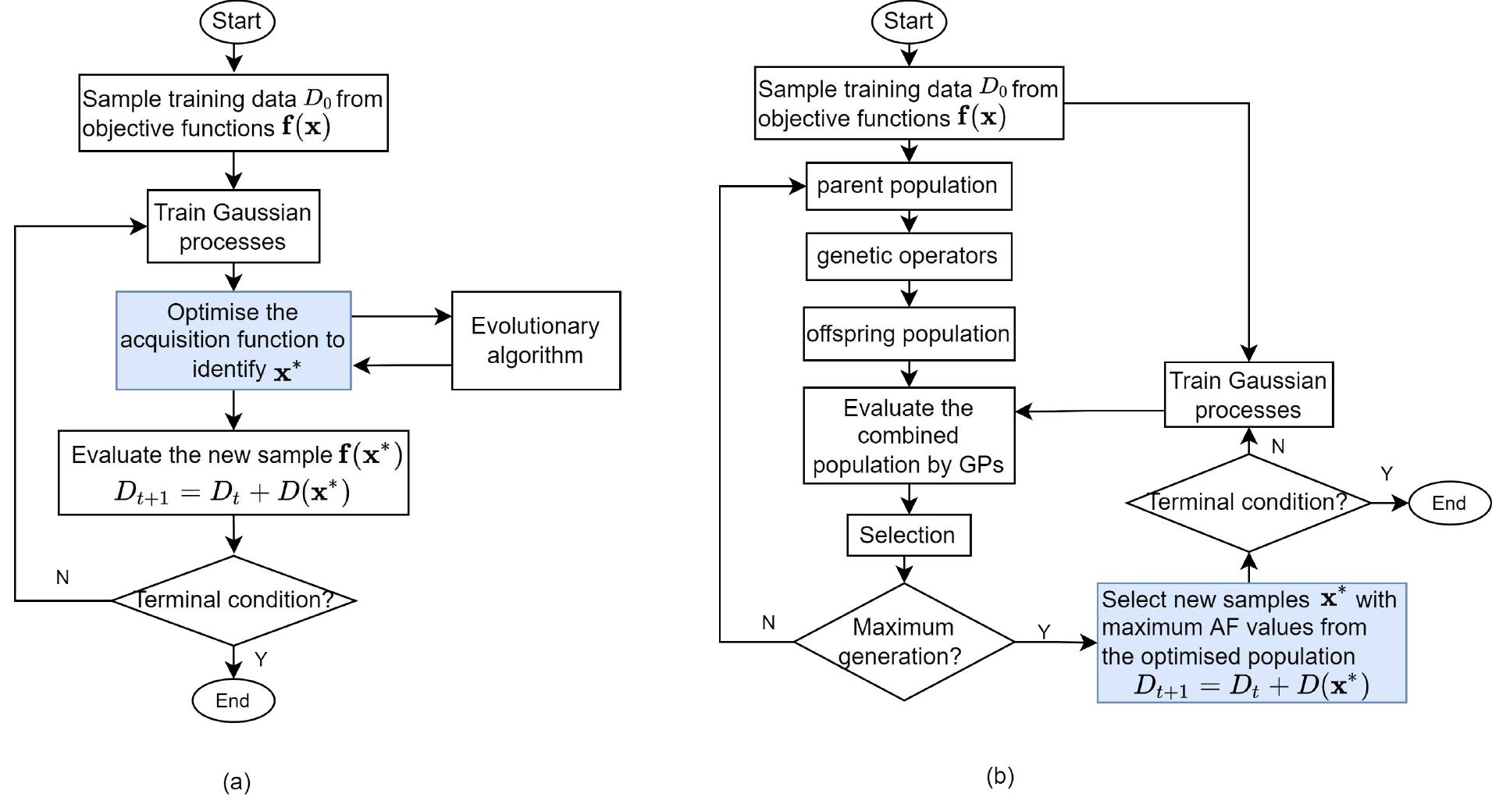}
\caption{Two main approaches combining between evolutionary algorithms with Bayesian optimization: (a) evolutionary Bayesian optimization, and (b) Bayesian evolutionary optimization. In (b), the fitness functions for environmental selection in the evolutionary algorithm may be different from the acquisition function in infilling samples.}
\label{beo_ebo}
\end{figure}

Since MOEAs have been successful in solving MOPs, it is straightforward to combine Bayesian optimization with MOEAs. This way, GPs and existing acquisition functions for single-objective optimization can be directly applied to each objective in MOPs. According to the way in which Bayesian optimization and evolutionary algorithms work together, the combinations can be further divided into two groups, evolutionary Bayesian optimization (EBO) and Bayesian evolutionary optimization (BEO) \cite{qin2019bayesian}. In EBO, as shown in Fig. \ref{beo_ebo} (a) Bayesian optimization is the basic framework in which the AF is optimized using an evolutionary algorithm. 
By contrast, in BEO, as shown in Fig. \ref{beo_ebo} (b), the evolutionary algorithm is the basic framework, where the AF is adopted as a criterion for selecting offspring individuals to be sampled. However, the objective functions in environmental selection of the MOEA may be different from the AFs.  

Many studies \cite{zhang2009expensive,chugh2016surrogate,knowles2006parego} have explored the applications of MOEAs with Gaussian processes as surrogate models for handling computationally expensive MOPs. The differences that distinguish these methods lie in the adopted MOEAs and the strategy for selecting new samples. Typically, decomposition based MOEAs use a scalarizing function, such as the Tchebycheff scalarizing function or the weighted sum, to generate a set of single-objective problems. ParEGO \cite{knowles2006parego} is an early work in this category: the augmented Tchebycheff function with a set of randomly generated weight vectors is adopted to construct multiple single-objective optimization problems, to which the traditional acquisition functions can be directly applied to identify new samples. In this way, only one new sample that maximizes the EI is evaluated at each iteration. It is desirable to develop multi-objective Bayesian optimization approaches that can produce several sample points at each iteration, which can be naturally achieved by MOEAs. 
By contrast, an MOP can be decomposed into multiple single-objective sub-problems, as done in the multiobjective evolutionary algorithm based on decomposition (MOEA/D) \cite{zhang2007moea} and the reference vector guided evolutionary algorithm (RVEA) \cite{cheng2016reference}. After that, Bayesian optimization can be applied to solve the sub-problems. For example, in MOEA/D-EGO \cite{zhang2009expensive}, Tchebycheff scalarizing function is used to decompose an MOP into a set of single-objective subproblems. Instead of constructing a model for each subproblem, MOEA/D-EGO divides the training samples into a number of subsets using a fuzzy clustering method, and subsequently a GP model is constructed for each cluster to reduce the computational cost. The EI is optimized and a set of new samples are selected from the population. Alternatively, Namura \emph{et al.} \cite{namura2017expected} adopted penalty-based boundary intersection (PBI) function to generate several single-objective problems. In the Kriging-assisted RVEA (K-RVEA) \cite{chugh2016surrogate}, a reference vector is used to decompose the MOP into a number of sub-problems. Then, the most uncertain solution is selected for sampling for each sub-problem if the diversity of the overall population needs to be promoted; otherwise, the solution having the best penalized angle distance according to the predicted objective values will be selected for each sub-problem. RVEA is also adopted as the optimizer in \cite{wang2020adaptive} to address expensive MOPs, where the predicted objective value and the uncertainty are weighted together as an acquisition function, and the weights are tuned to balance exploration and exploitation. 

Non-dominated sorting is another approach widely adopted in MOEAs. For example, Shinkyu \emph{et al} \cite{jeong2005efficient} proposed an extension of EGO using a non-dominated sorting based MOEA, called Multi-EGO. Multi-EGO maximizes the EIs for all objectives simultaneously, thus the non-dominated sorting is employed to select new samples. In a recent work \cite{belakaria2020uncertainty}, non-dominated sorting is used to select a cheap Pareto front based on the surrogate models and then identify the point with the highest degree of uncertainty for sampling. Similarly, multi-objective particle swarm optimization (MOPSO) using non-dominated sorting  is adopted in \cite{lv2019surrogate,li2019multi} in combination with Bayesian optimization. 
\subsubsection{Performance indicator based AFs} 
Performance indicators were originally developed to assess and compare the quality of solution sets (rather than a single solution) obtained by different algorithms \cite{zitzler2003performance}. Various quality indicators have been proposed, including inverted generational distance (IGD) \cite {zhou2006combining} and hypervolume (HV) \cite{zitzler1999multiobjective}. HV calculates the volume of the objective space dominated by a set of non-dominated solutions $\mathcal{P}$ and bounded by a reference point $\mathbf{r}$,
\begin{equation}
\operatorname{HV}(\mathcal{P})=VOL\left(\cup_{\mathbf{y} \in \mathcal{P}}[\mathbf{y}, \mathbf{r}]\right)
\end{equation}
where $VOL(\cdot)$ denotes the usual Lebesgue measure, $[\mathbf{y}, \mathbf{r}]$ represents the hyper-rectangle bounded by $y$ and $r$. Hence, algorithms achieving a larger HV value are better.

Interestingly, performance indicators can be incorporated into MOEAs in different manners. They can be adopted as an optimization criterion in the environmental selection \cite{xu2020polar} since they provide an alternative way to reduce an MOP into a single-objective problem. For this reason, various multi-objective Bayesian optimization methods with a performance indicator based AF have been developed, among which HV is the most commonly used performance indicator. An early work is $\mathcal{S}$-Metric-Selection-based efficient global optimization (SMS-EGO) \cite{ponweiser2008multiobjective}, which is based on the $\mathcal{S}$ metric or HV metric. In SMS-EGO, a Kriging model is built for each objective, then HV is optimized to select new samples, where the LCB is adopted to calculate the fitness values. Similarly, TSEMO \cite{bradford2018efficient} uses Thompson sampling on the GP posterior as an acquisition function, optimizes multiple objectives with NSGA-II, and then selects the next batch of samples by maximizing HV.

Indeed, the combination of the EI and HV, which is known as expected hypervolume improvement (EHVI), is more commonly seen in the context of expensive MOPs. Given the current PF approximation $\mathcal{P}$, the contribution of a non-dominated solution $\left(\mathbf{x},\mathbf{y}\right)$ to HV can be calculated by
\begin{equation}
I(\mathbf{y}, \mathcal{P})=HV(\mathcal{P} \cup\{\mathbf{y}\})-HV(\mathcal{P}),
\end{equation}
The EHVI quantifies the expectation of the HV over the non-dominated area. Hence, the generalized formulation of EHVI is formulated as
\begin{equation}
\mathrm{EHVI}(\mathbf{x})=\int_{\mathbb{R}^{m}} I(\mathbf{y}, \mathcal{P}) \prod_{i=1}^{m} \frac{1}{\sigma_{i}(\mathbf{x})} \phi\left(\frac{y_{i}(\mathbf{x})-\mu_{i}(\mathbf{x})}{\sigma_{i}(\mathbf{x})}\right) \mathrm{d} y_{i}(\mathbf{x}).
\end{equation}

EHVI was first introduced in \cite{emmerich2006single} to provide a scalar measure of improvement for prescreening solutions, and then became popular for handling expensive MOPs \cite{li2018modified,yang2019multi}. Wagner \emph{et al.} \cite{wagner2010expected} studied different AFs for MOPs, indicating that EHVI has desirable theoretical properties. The comparison between the EHVI with other criteria \cite{shimoyama2012comparison}, such as EI and estimation of objective values shows that EHVI maintains a good balance between the accuracy of surrogates and the exploration of the optimization. Despite the promising performance, the calculation of EHVI itself is computationally intensive due to the integral involved, limiting its application to MOPs/MaOPs. A variety of studies have been done to enhance the computation efficiency for EHVI. In \cite{emmerich2006single}, Monte Carlo integration is adopted to approximate the EHVI. Emmerich \emph{et al.} \cite{emmerich2011hypervolume} introduced a direct computation procedure for EHVI, which partitions the integration region into a set of interval boxes. However, the number of interval boxes scales at least exponentially with the number of Pareto solutions and objectives. In a follow-up work, Couckuyt \emph{et al.} \cite{couckuyt2014fast} introduced an efficient way by reducing the number of the interval boxes. Similar to EHVI, an HV-based PI is proposed in \cite{couckuyt2014fast}, which is defined by the product of the improvement function and the PI. More recently, an attempt to improve the computational efficiency of EHVI has been made \cite{li2018modified}, which adopted the concept of the local upper bounds in the hypervolume improvement. Given EHVI's differentiability, Yang \cite{yang2019multi} derived a gradient-based search algorithm for EHVI to speed up the optimization. 

Another commonly used indicator is based on distance, especially the Euclidean distance. Expected Euclidean distance improvement (EEuI) \cite{keane2006statistical} defines the product of the probability improvement function and an Euclidean distance-based improvement function for a closed-form expression of a bi-objective optimization problem. A fast calculation method for EEuI is proposed using the Walking Fish Group (WFG) algorithm \cite{couckuyt2014fast}. Alternatively, the maximin distance improvement is adopted as the improvement function in \cite{svenson2016multiobjective}. The Euclidean distance improvement, the maximin distance improvement and the hypervolume improvement are also reported in \cite{zhan2017expected} based on the expected improvement matrix.  

\subsubsection{Information theory based AFs}
Given the popularity of information theoretic approaches in the context of single-objective Bayesian optimization, it is not surprising that many information-based AFs for tackling expensive MOPs have been proposed. For example, predictive entropy search is adopted to address MOPs, called PESMO \cite{hernandez2016predictive}. However, optimizing PESMO is a non-trivial task: a set of approximations are performed; thus the accuracy and efficiency of PESMO can degrade. A subsequent work is the extension of the output-space-entropy based AF in the context of MOPs, known as MESMO \cite{belakaria2019max}. Empirical results show that MESMO is more efficient than the PESMO. As pointed out in \cite{suzuki2020multi}, MESMO fails to capture the trade-off relations among objectives for MOPs where no points in the PF are near the maximum of each objective. To fix this problem, Suzuki \emph{at al.} \cite{suzuki2020multi} proposed a Pareto-frontier entropy search (PFES) that considers the entire PF, in which the information gain is formulated as
\begin{equation}
I\left(\mathcal{F}^{*} ; \mathbf{y} \mid \mathcal{D}_{n}\right) \approx H\left[p\left(\mathbf{y} \mid \mathcal{D}_{n}\right)\right]-\mathbb{E}_{\mathcal{F}^{*}}\left[H\left[p\left(\mathbf{y} \mid \mathcal{D}_{n}, \mathbf{y} \preceq \mathcal{F}^{*}\right)\right]\right]
\end{equation}
where $\mathcal{F}^{*}$ is the Pareto front, $\mathbf{y} \preceq \mathcal{F}^{*}$ denotes $\mathbf{y}$ is dominated or equal to at least one point in $\mathcal{F}^{*}$.

\begin{figure}[ht]
\centering
\includegraphics[width=\columnwidth]{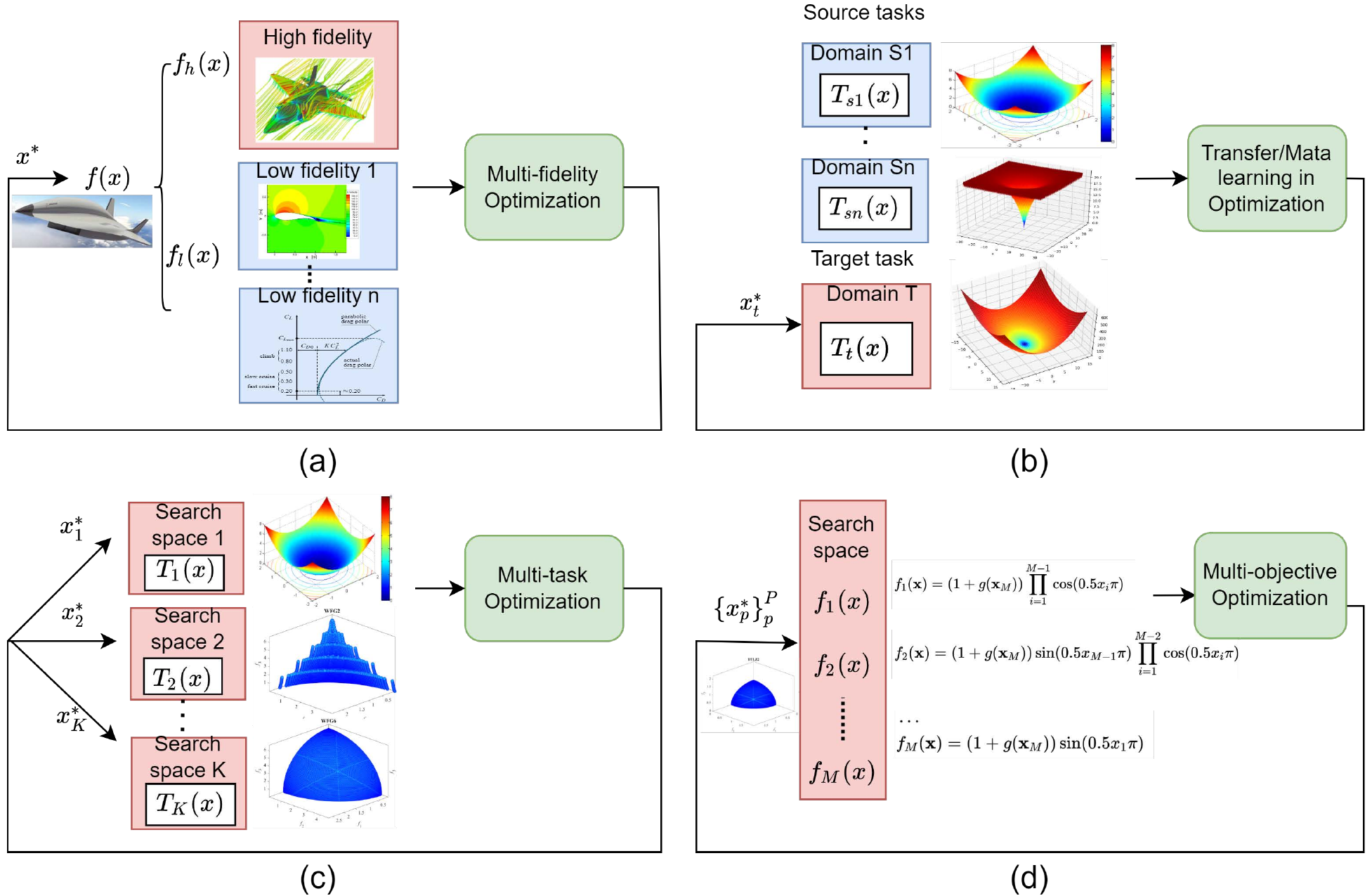}
\caption{The main difference between (a) multi-fidelity optimization, (b) transfer/meta learning in optimization, (c) multi-task optimization, and (d) multi-objective optimization. The target optimization task (denoted by red rectangles) in the four scenarios are different: while multi-objective optimization and multi-task optimization aim to effectively and concurrently optimize several problems, multi-fidelity optimization and transfer/meta learning in Bayesian optimization aim to accelerate the target optimization task by utilizing useful knowledge acquired from low fidelity simulations or similar source optimization tasks (denoted by blue rectangles). Note that in multi-task optimization, all tasks are equally important and knowledge transfer occurs between any of the related tasks. Finally, the difference between multi-objective optimization and multi-task optimization is that the former handles conflicting objectives of the same task, while each task in the latter can be a single/multi-objective problem.}
\label{DifferentProblem}
\end{figure}

\subsection{Multi-task Optimization}
Many black-box optimization problems are not one-off tasks. Instead, several related instances of the tasks can be simultaneously tackled, which is known as multi-task optimization. Suppose there are $K$ optimization tasks, $i=\{1,2, \ldots, K\}$, to be accomplished. Specifically, denote $T_i$ as the $i$-th task to be optimized and $X_i$ as the search space of $T_i$. Without loss of generality, assuming each task is a minimization problem, and multi-task optimization (MTO) aims to find a set of solutions $\left\{\mathbf{x}_{1}^{*}, \ldots, \mathbf{x}_{K}^{*}\right\}$ satisfying

\begin{equation}
\mathbf{x}_{i}^{*}=\underset{\mathbf{x} \in X_{i}}{\arg \min } T_{i}(\mathbf{x}), i=1,2, \ldots, K.
\end{equation}

There exist some conceptual similarities and overlaps between multi-task optimization and some other terms, such as multi-objective optimization, multi-fidelity optimization and transfer/meta learning. Similarities and differences are illustrated in Fig. \ref{DifferentProblem}. Note that the goal of multi-objective optimization is to handle conflicting objectives of the same task and find optimal tradeoff solutions. By contrast, MTO aims to effectively and concurrently optimize multiple tasks by leveraging the correlated information among different tasks, with each task in MTO being either a single- or multi-objective problem. While multi-fidelity optimization and transfer/meta learning focus on the target task, MTO treats all tasks equal and the knowledge transfer occurs between any related tasks. 
A detailed description of the differences between these optimization problems can be found in \cite{zhuang2020comprehensive}.

Multi-task Bayesian optimization (MTBO) aims to optimize a collection of related tasks at the same time, thereby speeding up the optimization process by taking advantage of the common information across the tasks. There are two requirements to achieve this. First, surrogate models that can learn the transferable knowledge between the tasks should be built. Second, the acquisition function should consider not only the exploration-exploitation balance, but also the correlation between the tasks, so that the data efficiency of optimization can be further improved by transferring knowledge between the related tasks. In the following, we present Bayesian optimization algorithms in which multi-task Gaussian models are constructed and specific acquisition functions are designed for MTO. 
The need for multi-task learning is ubiquitous across many applications in various fields, such as hyperparameter optimization of machine learning models \cite{swersky2013multi}, robotic manipulator inverse dynamics \cite{durichen2014multitask}, biomedical engineering, and biomedical engineering \cite{durichen2014multitask}. Hence, MLO has drawn considerable attention in the machine learning community, and many MLO models and applications have been explored. Among them, Gaussian process models have been extensively applied to learning a set of tasks on different data sets. More recently, BO has been applied to multi-task learning. In the following, the existing proposed multi-task Bayesian models and Bayesian optimization algorithms for MTL are presented, respectively.

\subsubsection{Multi-task Gaussian process} MTO benefits from transferring knowledge across different tasks assuming that the tasks are related to a certain degree. In the geostatistics community, the linear model of coregionalization (LMC) expresses the outputs as linear combinations of $Q$ independent random functions,
\begin{equation}
T_{i}(\mathbf{x})=\sum_{q=1}^{Q} a_{i, q} u_{q}(\mathbf{x}),
\end{equation}
where the latent function $u_{q}(\mathbf{x})$ is assumed to be a zero-mean Gaussian process with covariance as $k_{q}\left(\mathbf{X}, \mathbf{X}^{\prime}\right)$, and $a_{i, q}$ is the coefficient for $u_{q}(\mathbf{x})$.
In the context of machine learning, many Bayesian multi-task models can be viewed as variations of the LMC with different parameterizations and constraints. A representative work is called multi-task GP \cite{williams2007multi}, which uses the intrinsic coregionalization model kernel. Besides the covariance function over inputs $k_{\mathcal{X}}\left(\mathbf{x}, \mathbf{x}^{\prime}\right)$, a task covariance matrix $k_{\mathcal{T}}\left(t, t^{\prime}\right)$ is introduced as coregionalization metrics to model the inter-task similarities. Consequently, the product kernel can be derived as follows:
\begin{equation}
k\left((\mathbf{x}, t),\left(\mathbf{x}^{\prime}, t^{\prime}\right)\right)=k_{\mathcal{X}}\left(\mathbf{x}, \mathbf{x}^{\prime}\right)\otimes k_{\mathcal{T}}\left(t, t^{\prime}\right)
\label{MTGP}
\end{equation}
where $\otimes$ denotes the Kronecker product, and $t,t^{\prime} \in \mathcal{T}$,  $k_{\mathcal{T}}\left(t, t^{\prime}\right)$ is a positive semi-definite matrix, which is guaranteed by the Cholesky decomposition. The multi-task GP suffers from a high computational complexity of $O(Tn^3)$. To improve the scalability of MTGP, an efficient learning algorithm using self-measuring similarity is introduced to construct the covariance matrics in \cite{hayashi2012self}.

In LMC models, the correlated process is expressed by a linear combination of a set of independent processes, called instantaneous mixing. Such a method is limited to scenarios where one output process is a blurred version of the other. Alternatively, convolution processes are employed to account for correlations across outputs, and each output can be expressed through a convolution integral between a smoothing kernel and a latent function \cite{alvarez2011computationally}. However, the approach is criticized for its computational and storage complexity. 

For example, Bakker and Heskes \cite{bakker2003task} proposed a Bayesian neural network for MTL in a hierarchical Bayesian perspective. While input-to-hidden weights are shared by all tasks, hidden-to-output weights are task-dependent by placing a prior distribution on the model parameters. Following the similar idea, a more general GP with parametric covariance functions is introduced by Lawrence and Platt \cite{lawrence2004learning} for MTL and knowledge sharing. Moreover, the informative vector machine is adopted to reduce computation by sparsifying the covariance matrix. Instead of learning the covariance matrix in a parameteric manner, the use of hierarchical Bayesian modeling on GPs is presented by Yu \emph{et al.} \cite{yu2005learning} and Schwaighofer \emph{et al.} \cite{schwaighofer2005learning}, using a normal-inverse Wishart prior distribution for the mean and covariance function. The assumption behind sharing the same prior over the mean and the covariance matrix is that all tasks are correlated. However, such an assumption may not hold. As a treatment to the outlier tasks, Yu \emph{et al.} \cite{yu2007robust} presented a robust extension of the previous studies \cite{yu2005learning,schwaighofer2005learning} by using heavier tailed t-Processes. To facilitate efficient inference of the work \cite{yu2005learning}, pseudo inputs are adopted to derive a sparse construction for the GP \cite{lian2015multitask}. As stated in \cite{zhang2021survey}, MTL can boost the performance of reinforcement learning, coined as multi-task reinforcement learning (MRL). Few attempts have been done in this line of research, and some of them has revisited BO. Ghavamzadeh (2010) exploit shared structure in the value functions between related MDPs. However, their approach is designed for on-policy multi-task policy evaluation, rather than computing optimal policies.  
\subsubsection{Acquisition functions in MTO} Although many attempts have been made to propose multi-task models, only recently a few multi-task Bayesian optimization algorithms have been proposed, especially in the field of hyperparameter optimization in machine learning. Swersky and Snoek \cite{swersky2013multi} extend the multi-task GP \cite{williams2007multi} to Bayesian optimization for knowledge transfer in tuning hyperparameters, where a new AF based on entropy search is proposed by taking cost into consideration. Similar ideas that adopt multi-task GPs or design a new AF introducing a trade-off between information gain and cost minimization can be found in \cite{moss2020mumbo} and \cite{klein2017fast}. Bardenet \emph{et al.} \cite{bardenet2013collaborative} considered the hyper-parameter optimization for deep belief networks with different features of the dataset, and proposed collaborative tuning of several problems. While a GP is used to predict the algorithm's performance, each dataset is visited at each iteration and a new sample is selected by maximizing EI on that dataset.

In contextual policy search (CPS), a joint GP model over the context-parameter space is learned, allowing knowledge acquired from one context to be generalized to similar contexts. ES has been extended to CPS \cite{metzen2015active} by averaging the expected entropy at different points in a set of randomly sampled contexts. Unfortunately, the performance of sampling-based entropy search is not competitive, and its performance deteriorates in the presence of outliers. Hence, Metzen \cite{metzen2016minimum} further investigated minimum regret search to explicitly minimize the expected simple regret. More recently, Thompson sampling has been extended to multi-task optimization by sampling from the posterior to identify the next task and action \cite{char2019offline}, which is theoretically guaranteed. 
Metzen \emph{et al.} \cite{metzen2015bayesian} proposed a Bayesian optimization approach (BO-CPS) to handle CPS and adopted the GP-UCB to select parameters for a given context. The global maximizer DIRECT \cite{jones1993lipschitzian} is adopted to optimize the GP-UCB.
\subsection{Multi-fidelity optimization}
Bayesian optimization generally assumes that only the target expensive objective function is available, which is referred to as single-fidelity optimization. In many practical problems, however, the evaluation of the target function $f(\mathbf{x})$ can often be run at multiple levels of fidelity with varying costs, $\left\{f_{1}(\mathbf{x}), \ldots, f_{M}(\mathbf{x})\right\}$, where the higher the fidelity $m \in\{1,2, \ldots, M\}$, the more accurate but costly the evaluation will be. For example, in the optimization of the ship hull shape, both low-fidelity inexpensive and high-fidelity expensive hydrodynamic models can be used. 
This is known as multi-fidelity optimization (MFO), which can be seen as a subclass of multi-task learning, where the group of related functions can be meaningfully ordered by their similarity to the objective function.

MFO aims to accelerate the optimization of the target objective and reduce the optimization cost by jointly learning the maximum amount of information from all fidelity models. To achieve this, Bayesian optimization undertakes two changes to make use of multiple fidelity data, namely multi-fidelity modeling and a new sample selection, which will be discussed in detailed in the following.
\subsubsection{Multi-fidelity models}
Typically, multi-fidelity Bayesian optimization builds surrogate models of different levels of fidelity either by learning an independent GP for each fidelity \cite{kandasamy2016gaussian}, or jointly modeling multi-fidelity data to capture the correlation between the different fidelity data, such as multi-output GP and deep neural networks. Among them, one most popular multi-fidelity model is Co-Kriging \cite{myers1982matrix}. Kennedy and O'Hagan \cite{kennedy2000predicting} proposed an autoregressive model to approximate the expensive high-fidelity simulation $\hat{y}_{H}(\mathbf{x})$ by the sum of the low-fidelity Kriging model $\hat{y}_{L}(\mathbf{x})$ and a discrepancy model $\hat{\delta}(\mathbf{x})$, formulated as 
\begin{equation}
\hat{y}_{H}(\mathbf{x})=\rho \hat{y}_{L}(\mathbf{x})+\hat{\delta}(\mathbf{x})
\label{MFO}
\end{equation}
where $\rho$ denotes a scaling factor minimizing the discrepancy between $\rho \hat{y}_{L}(\mathbf{x})$ and high-fidelity model at the common sampling points. Thus, high-fidelity model can be enhanced by acquiring information from the low-fidelity inexpensive data. 
Following this basic idea, Forrester \emph{et al.} \cite{forrester2007multi} and Huang \emph{et al.} \cite{huang2006sequential} have further investigated the extensions of Co-Kriging to MOPs. Later, a Bayesian hierarchical GP model is developed in \cite{qian2008bayesian} to account for complex scale changes from low fidelity to high fidelity. To improve the computational efficiency, a recursive formulation for Co-Kriging was proposed in \cite{le2014recursive}, assuming that the training datasets for $\hat{y}_{H}(\mathbf{x})$ and $\hat{y}_{L}(\mathbf{x})$ have a nested structure, i.e., the training data for the higher fidelity levels is a subset of that of a lower fidelity level. Hence, the GP prior $\hat{y}_{L}(\mathbf{x})$ in Eq. \ref{MFO} is replaced by the corresponding GP posterior, improving the efficiency of the hyperparameter estimations. Following this idea, the autoregressive multi-fidelity model given in Eq. \ref{MFO} has been generalized by replacing the scaling factor $\rho$ with a non-linear mapping function \cite{perdikaris2017nonlinear}.

The multi-fidelity Kriging model has been employed in many domains of research including aerodynamics \cite{brevault2020overview,bailly2019multifidelity}, engineering design \cite{keane2012cokriging,toal2014multifidelity}, bandit optimisation problems \cite{song2019general,sen2018multi,kandasamy2017multi}, multi-objective optimization problems \cite{belakaria2020multi} and hyperparameter tuning \cite{sen2018multi}. It is worthy of noting that multi-fidelity optimization for bandit problems and MOPs typically focuses on the design of new AFs, which we will present in the following. 

\subsubsection{Acquisition functions for multi-task optimization}
Based on multi-task models \cite{le2014recursive,kennedy2000predicting,forrester2007multi}, the design of sophisticated AFs to select both the input locations and the fidelity in the MFO setting has attracted much research interest. Earlier multi-fidelity AFs focused on the adaptation of EI. Huang \emph{et al.} \cite{huang2006sequential} proposed an augmented EI function to account for different fidelity levels of an infill point. Specifically, the proposed EI is the product of the expectation term, the correlation between the LF and HF models, the ratio of the reduction in the posterior standard deviation after a new replicate is added \cite{huang2006global}, and the ratio between the evaluation cost of the LF and HF models. To enhance the exploration capability of augmented EI, Liu \emph{et al.} \cite{liu2018sequential} proposed a sample density function that quantifies the distance between the inputs to avoid clustered samples. On the other hand, a new formulation of uncertainty is introduced to the EI in the context of MFO in \cite{zhang2018variable}. A recent work develops an adaptive EI to select updating samples, so that a different EI function is used based on which fidelity to query \cite{hao2020adaptive}.A closely related AF to the EI, the KG, also has been applied to the MFP \cite{wu2020practical}.

UCB has been widely used in MFO, especially in bandit problems. An early work on principled AF based UCB for MFO is MF-GP-UCB \cite{kandasamy2016gaussian}. The MF-GP-UCB algorithm first formulates an upper bound for each fidelity, among which the minimum bound is identified to be maximized for selecting the new sample. Having selected the new point, a threshold is introduced to decide which fidelity to query. In a follow-up work \cite{kandasamy2017multi}, MF-GP-UCB is extended to the continuous fidelity space. Sen \emph{et al.} \cite{sen2018multi} developed an algorithm based on a hierarchical tree-like partitioning, and employed MF-GP-UCB to select the leaves. The motivation behind this method is to explore coarser partitions at lower fidelities and proceed to finer partitions at higher fidelities when the uncertainty has shrunk. Following this idea, Kandasamy \emph{et al.} \cite{kandasamy2019multi} adopted MF-GP-UCB to explore the search space at lower fidelities, and then exploit the high fidelities in successively smaller regions. Another plausible way to address bandit problems in multi-fidelity settings is information-based methods. A multi-fidelity mutual-information greedy optimization (MF-MI-Greedy) is introduced in \cite{song2019general}. Each round of MF-MI-Greedy includes an exploration phase to explore the low fidelity actions and an optimization phase to optimize the payoff function at the target fidelity.

Recently, information-theoretic approaches have become popular in  MFO. For example, ES with the Co-Kriging model is adopted in \cite{marco2017virtual} to solve a two-fidelity optimization. McLeod \emph{et al.} \cite{mcleod2017practical} introduced an environmental variable to denote the varying fidelities, thus a GP can be built on the augmented space. Then, PES is adopted as the AF and a fast sampling strategy is employed to reduce the computational cost. In \cite{zhang2017information}, unknown functions with varying fidelities are jointly modeled as a convolved Gaussian process \cite{alvarez2011computationally}, then a multi-output random feature approximation is introduced to calculate PES. Since it is non-trivial to calculate the multi-fidelity AFs based on ES/PES, MES has been extended to MFO due to its high computational efficiency \cite{takeno2020multi}. 

\subsection{Transfer/Meta Learning}
Although Bayesian optimization offers a powerful data-efficient approach to global black-box optimization problems, it considers each task separately and often starts a search from scratch, which needs a sufficient number of expensive evaluations before achieving high-performance solutions. To combat such a "cold start" issue, transfer/meta learning in Bayesian optimization has attracted a surge of interest in recent years. Given a set of auxiliary/source domains $D_s$ and optimization tasks $T_s$, a target domain $D_T$ and optimization task $T_T$, transfer/meta learning in Bayesian optimization aims to leverage knowledge from previous related tasks $T_s$ to speed up the optimization for the target task $T_T$. A well-studied example is hyperparameter optimization of a machine learning algorithm on a new dataset (target) with observed hyperparameter performances on the other datasets (source/meta-data). The availability of meta-data from previous related tasks in hyperparameter optimization has motivated a simple strategy to speed up the optimization process on a new dataset, called meta-initialization. The core idea behind this is to initialize a hyperparameter search based on the best hyperparameter configurations for similar datasets \cite{feurer2015initializing}. Typically, the two terms, i.e., transfer/meta learning, are used interchangeably in the context of Bayesian optimization. Note that in the Bayesian optimization community, knowledge transfer has also been investigated under the several umbrellas, including multi-task learning and multi-fidelity optimization, which may overlap with the broad field of transfer learning. 

Intuitively, the optimization of target task may suffer from negative transfer if the learned knowledge degrades the performance. Hence, the success of transfer learning is heavily conditioned on the similarity between source and target tasks. According to the method for capturing the similarity, we classify the Bayesian optimization algorithms coupled with transfer learning techniques into the following three groups. 
The difference between transfer/meta learning and the other notions in the context of BO lies in the problem setup: multi-task learning aims to optimise all tasks simultaneously by allowing knowledge to transfer among them. Multi-fidelity optimization has access to low-fidelity evaluations (source) during the optimization process.
\subsubsection{Meta-initialization}
The availability of meta-data from previous related tasks in hyperparameter optimization has motivated a simple strategy to speed up the optimization process on a new dataset, called meta-initialization. The core idea behind this is to initial a hyperparameter search based on the best hyperparameter configurations for similar datasets. To achieve this, Feurer \emph{et al.} \cite{feurer2015initializing} introduced a negative Spearman correlation coefficient to measure the similarity between different datasets, while Wistuba \emph{et al.} \cite{wistuba2015learning} identified the initial hyperparameter configurations via optimising a meta-loss.

\subsubsection{Hierarchical model}
Hierarchical models learned across the entire datasets arise as a natural solution to making use of the knowledge from related source domains \cite{tighineanu2021transfer}. For example, Bardenet \emph{et al.} \cite{bardenet2013collaborative} noted that the loss values on different datasets may differ in scale, motivating a ranking surrogate to map observations from all runs into the same scale. However, this approach suffers from a high computational complexity incurred by the ranking algorithm. To address this problem, Yogatama and Mann \cite{yogatama2014efficient} suggested to reconstruct the response values by subtracting the per-dataset mean and scaling through the standard deviation, while Golovin \emph{et al.} \cite{golovin2017google} proposed an efficient hierarchical GP model using the source posterior mean as the prior mean for the target.

\subsubsection{Multi-task Gaussian process}
Since multi-task GP models are powerful for capturing the similarity between the source and target tasks, Swersky \emph{et al.} \cite{swersky2013multi} conducted a straightforward knowledge transfer using a multi-task GP. Meanwhile, the semi-definite (PSD) matrix in multi-task GPs (see Eq. \ref{MTGP}) has been modified to improve the computational efficiency \cite{yogatama2014efficient, min2020generalizing}. On the other hand, Joy \emph{et al.} \cite{joy2019flexible} assumed that the source data are noisy observations of the target task, so that the difference between the source and target can be modeled by noise variances. Following this idea, Ramachandran \emph{et al.}  \cite{ramachandran2018selecting} further improved the efficiency of the knowledge transfer by using a multi-bandit algorithm to identify the optimal source.

\subsubsection{Weighted combination of GPs}
Knowledge transfer in Bayesian optimization can also be achieved by a weighted combination of GPs. Instead of training a single surrogate model on a large training data set (i.e., the historical data), Schilling \emph{et al.} \cite{schilling2016scalable} suggested to use the product of GP experts to improve the learning performance. Specifically, an individual GP is learned on each distinct dataset. This way, the prediction on a target data provided by the product of the individual GPs is a sum of means with weights adjusted with regard to the GP uncertainty. Different strategies have been proposed to adapt the weights in the combination \cite{da2019fast,feurer2018scalable}. In multi-objective optimization, Min \emph{et al.} \cite{min2017multiproblem} proposed to identify the weights by optimizing the squared error of out-of-sample predictions. Interestingly, in \cite{ramachandran2018information} the location of the global optimum for the target is modeled by combining the distribution of the optimum of each source task. The weight in the mixture distribution is proportional to the similarity between the source and target, which is measured by Kullback-Leibler divergence. 

In a complementary direction, a few attempts have been dedicated to leveraging the meta-data within the acquisition function in a similar fashion to the weighted combination of GPs. A representative work is called transfer AF (TAF) \cite{wistuba2018scalable}, which is defined by the weighted average of the expected improvement on the target dataset and source datasets. More recently, Volpp \emph{et al.} \cite{volpp2020meta} adopted reinforcement learning to achieve this. 

\subsection{Parallel/Batch Bayesian optimization}
The canonical Bayesian optimization is inherently a sequential process since one new data is sampled in each iteration, which might be inefficient in many applications where multiple data points can be sampled in parallel \cite{nguyen2021optimal}. A strength of sequential Bayesian optimization is that a new data point is selected using the maximum available information owing to the immediately updated GP, and therefore searching for multiple query points simultaneously is  more challenging. With the growing availability of parallel computing, an increasing number of studies exploring batch Bayesian optimization have been carried out, which can be roughly classified into two groups. One is the extension of the existing AFs to batch selection, and the other is problem reformulation.  

\subsubsection{Extensions of the existing AFs}
A pioneering multi-points acquisition function is the parallelized version of the expected improvement (EI), called q-points EI (q-EI) \cite{ginsbourger2008multi,ginsbourger2010kriging}. The q-EI is straightforwardly defined as the expected improvement of the $q$ points beyond the current best observation. However, the exact calculation of q-EI depends on the integral of q-dimensional Gaussian density, and therefore becomes intractable and intensive as $q$ increases. Hence, Ginsbourger \emph{et al.} \cite{ginsbourger2008multi} sequentially identified $q$ points by using Kriging believer or constant liar strategies to replace the unknown output at the last selected point, facilitating the batch selection based on q-EI. Treatments for the intractable calculation of q-EI have been investigated in \cite{ginsbourger2010kriging,wang2020parallel,chevalier2013fast}. Besides, an asynchronous version of q-EI is presented in \cite{janusevskis2012expected}.

The parallel extension of the GP-UCB has been widely investigated owing to its theoretical guarantees, i.e., the sublinear growth of cumulative regret. An extension of GP-UCB, called GP-BUCB, is proposed to leverage the updated variance, encouraging more exploration \cite{desautels2014parallelizing}. Interestingly, the GP-BUCB has been generalized to a multi-agent distributed setting \cite{daxberger2017distributed}.
Similarly, a GP-UCB approach with pure exploration (GP-UCB-PE) is proposed in \cite{contal2013parallel}, which identifies the first query point via the GP-UCB, while the remaining ones are selected by maximizing the updated variance. Since MOEAs can provide a set of non-dominated recommendations, they are well-suited for determining the remaining points by simultaneously optimizing the predicted mean and variance \cite{gupta2018exploiting}. In addition, distance exploration can be used to achieve this, avoiding selecting the same query points in a batch \cite{nguyen2018practical}. Both GP-BUCB and GP-UCB-PE greedily collect new points by maximizing the information gain estimated by the posterior variance. More diverse batches can be probed by sampling from determinantal point processes (DPPs) \cite{kathuria2016batched,wang2017batched}. Similarly, a variant of DPPs, called k-DPPs, is adopted to select a batch of neural network architectures for parallel evaluations \cite{nguyen2021optimal}.

With the rapidly growing interest in batch Bayesian optimization, more AFs have been extended to the parallel setting. For example, parallelized PES  \cite{shah2015parallel} and KG (q-KG) \cite{wu2016parallel} are developed to jointly identify a batch of points to probe in the next iteration, rendering, however, a poor scalability to the batch size. Interestingly, a state-of-the-art information-based AF, called trusted-maximizers entropy search (TES), is proposed by introducing trusted maximizers to simplify the information measure \cite{nguyen2021trusted}, which is well scalable to the batch size. TS can also be extended to the parallel setting by sampling $q$ functions instead \cite{hernandez2016distributed}. More recently, TS has attracted much attention, as the inherent randomness of TS automatically achieves a balance between exploitation and exploration \cite{kandasamy2018parallelised}. Similarly, it is suggested to sample from a probability distribution over an AF defined by the GP's hyperparameters \cite{de2019sampling} , while in \cite{de2021asynchronous},  TS is combined with the $\epsilon$-greedy acquisition function to account for asynchronous parallel optimization problems \cite{de2021asynchronous}. Note that the performance of TS is not necessarily better than traditional AFs, such as EI and UCB.

\subsubsection{Problem reformulation}
Much effort has been devoted to developing new batch approaches by reformulating the optimization problem of AFs in parallel Bayesian optimization. One interesting direction aims to develop new batch AFs to select input batches that closely match the expected recommendation of sequential methods. For example, a batch objective function minimizing the loss between the sequential selection and the batch is defined in \cite{azimi2010batch}, which corresponds to a weighted k-means clustering problem. Given that the sequentially selected inputs are sufficiently different from each other, a maximization-penalization strategy is introduced by adding a local penalty to the AF \cite{gonzalez2016batch}. Liu \emph{et al.} \cite{liu2021batch} applied a multi-start strategy and gradient-based optimizer to optimize the AF, aiming to identify the local maxima of the AF. In addition, the multi-objective optimizer is a promising approach to finding a batch of query points \cite{lyu2018batch,DBLP:journals/tcad/ZhangYYZZ22}, particularly for addressing expensive MOPs \cite{chugh2016surrogate,wang2020adaptive}. Similarly, sequentially optimizing multiple AFs is amenable to generating batches of query points \cite{joy2020batch}. To better balance exploration and exploitation, different selection metrics can be combined \cite{gong2019quantile,hu2018parallelizable}. Moreover, in \cite{wang2018batched, young2020distributed} local GPs are constructed so that batches of new samples that correspond to each GP can be collected. 

\section{Challenges and Future Directions}
\label{future}
Bayesian optimization is a well-established powerful optimization method for handling expensive black-box problems, which has found many successful real-world applications. Despite all these advances, numerous challenges remain open. In fact, the field of Bayesian optimization keeps very active and dynamic, partly because an increasing number of new applications in science and technology poses new challenges and demands. In the following, we present several most recent important developments in Bayesian optimization and discuss future research directions. 

\subsection{Distributed Bayesian optimization}
Distributed optimization problems are commonly seen in the real world. Despite a proliferation of studies on parallel or batch Bayesian optimization in recent years, most of them require a central server to construct a single surrogate model with few exceptions. For example, a straightforward distributed Bayesian optimization, called HyperSpace, has been proposed by Young \emph{et al.} \cite{young2018hyperspace,young2020distributed} for hyperparameter optimization. HyperSpace partitions the large search space with a degree of overlap and all possible combinations of these hyperspaces are generated and equipped with a GP model, allowing us to run the optimization loop in parallel. Thompson sampling can be fully distributed and handle the asynchronously parallel setting \cite{hernandez2017parallel}, although it fails to perform well due to its inherent randomness. Barcos and Cantin \cite{garcia2019fully} presented an interpretation of Bayesian optimization from the Markov decision process perspective and adopted Boltzmann/Gibbs policy to select the next query, which can be performed in a fully distributed manner.

Several questions remain open in design of distributed Bayesian optimization. First, it is of fundamental importance to achieve a trade-off between the convergence rate and communication cost. The convergence of distributed Bayesian optimization needs more rigorous theoretical proof and requires further improvement, and the computational gains will be offset in the presence of communication latencies. Second, it is still barely studied how to handle asynchronous settings that result from time-varying communication costs, different computation capabilities and heterogeneous evaluation times. Third, it is an important yet challenging future direction to take more practical scenarios into consideration, such as complex communication networks and communication constraints. 
 

\subsection{Federated Bayesian optimization}
While the rapidly growing sensing, storage and computational capability of edge devices has made it possible to train powerful deep models, increasing concern over data privacy has motivated a privacy-preserving decentralized learning paradigm, called federated learning \cite{mcmahan2017communication}. The basic idea in federated learning  is that the raw data remains on each client, while models trained on the local data are uploaded to a server to be aggregated, thereby preserving the data privacy. Adapting Bayesian optimization to the federated learning setting is motivated by the presence of black-box expensive machine learning and optimization problems.

Dai \emph{et al.} \cite{dai2020federated} explored the application of Bayesian optimization in the horizontal federated learning setting, where all agents share the same set of features and their objective functions are defined on a same domain. Federated Thompson sampling (FTS), which samples from the current GP posterior on the server with a probability of $p$ and consequently samples from the GP provided by the clients with a probability $1-p$. However, FTS lacks a rigorous privacy guarantee. To remedy this drawback, differential privacy \cite{dp2008}, a mathematically rigorous approach to privacy preservation, is introduced into FTS, called DP-FTS \cite{dai2021differentially}. 
More specifically, the DP-FTS partitions the search space into disjoint sub-spaces and then equips each sub-space with an agent. Instead of setting a target agent, DP-FTS adds a central server to perform the DP strategy. After aggregating the model and broadcasting to all agents by the server, the TS is performed on each agent to select the new query.  

Instead of using GPs as surrogates, Xu \emph{et al.} \cite{xu2021federated1} proposed to use radial-basis-function networks (RBFNs) on local clients. A sorting averaging strategy is proposed to construct a global surrogate on the server, where each local RBFN is sorted by a matching metric, and the parameters of each local surrogate are averaged according to the sorted index. To identify new samples, the local and global surrogates work together to provide a mean and variance predictions, and a federated LCB is adopted as an AF. The RBFN-based federated optimization was extended to handle multi/many-objective optimization problems \cite{xu2021federated2}. 

Although much work addressing challenges in federated learning, including communication efficiency, systems and data heterogeneity, and privacy protection have been reported, privacy-preserving optimization brings with many new questions. First, since GP is non-parameter models, it cannot be directly applied to the federated setting. One idea is to approximate the GP model with random Fourier feature approximates \cite{dai2020federated}, in which representative power and computation efficiency should be taken into consideration. Second, Thompson sampling is adopted as AF due to its ability to handle heterogeneous settings; however, it is criticized by its poor performance compared with other AFs. Hence, further investigation in new acquisition method is an interesting yet challenging research direction. Finally, privacy protection in federated Bayesian optimization remains elusive, and more rigorous definitions of threat models in the context of distributed optimization is highly demanded.

\subsection{Dynamic optimization}
In many real-world applications, such as network resource allocation, recommendation systems, and object tracking, the objective function to be optimized may change over time. Such optimization scenarios are known as dynamic optimization or time-dependent problems. Solving such problems are challenging for most optimization techniques designed for stationary problems \cite{yazdani2021}. Although various Bayesian optimization algorithms for solving static expensive black-box problems have been proposed, only a few methods have been developed to handle dynamic optimization problems. 

Most Bayesian optimization methods for dynamic optimization rely on the multi-armed bandit setting with time-varying reward functions. Bogunovic \emph{et al.} \cite{bogunovic2016time} introduced a simple Markov model for the reward functions using GPs, allowing the GP model to vary at a steady rate. Instead of treating all the samples equally important, \emph{resetting} \cite{zhao2020simple}, \emph{temporal kernel} \cite{chen2021transfer}, \emph{sliding window} \cite{zhou2021no}, and \emph{weighted GP model} \cite{deng2021weighted} have been proposed to achieve forgetting-remembering trade-off. More recently, a time-dependent objective is optimized at a given future time combined with a two-step look-ahead AF \cite{renganathan2021lookahead}. Nevertheless, the construction of effective surrogates for time-dependent objective functions, the design of acquisition functions to identify promising solutions and track the optimum remain challenging problems. Moreover, it is interesting to incorporate advances in machine learning, such as transfer learning, for leveraging informative from the previous runs. 


\subsection{Heterogeneous evaluations} 
Bayesian optimization implicitly assumes that the evaluation cost in different regions of the search space is the same. This assumption, however, can be violated in practice. For example, the evaluation times of different hyperarameter settings and the financial cost for steel or drug design using different ingredients \cite{abdolshah2019cost} may vary dramatically. Moreover, in multi-objective optimization, different objectives may have significantly different computational complexities, known as heterogeneous objective functions \cite{allmendinger2015multiobjective}. Handling heterogeneous evaluation costs that arise in both search spaces and objective spaces has attracted increased attention, motivating the development of cost-aware Bayesian optimization. 

Most cost-aware Bayesian optimization algorithms focus on single-objective optimization problems. Snoek \emph{et al.} \cite{snoek2012practical} introduces an AF called \emph{expected improvement per second} (EIps) to balance between the cost efficiency and evaluation quality via dividing EI by cost. This approach, however, tends to exhibit good performance only when the optimal solution is computationally cheap. To remedy this drawback, a cost-cooling strategy in a cost apportioned Bayesian optimization (CArBO) \cite{lee2020cost} de-emphasizes the heterogeneous costs as the optimization proceeds. Besides, CArBO conducts a cost-effective initialization to achieve a set of cheap and well-distributed initial points, aiming to explore cheaper areas first. In \cite{lee2021nonmyopic}, an optimization problem constrained by a cost budget is formulated as a constrained Markov decision process and then a rollout AF with a number of look-ahead steps is proposed. The cost-aware Bayesian optimization has also been extended to multi-objective problems where the evaluation costs are non-uniform in the search space \cite{abdolshah2019cost}.

To handle heterogeneous computational costs of different objectives in multi-objective optimization, simple \emph{Interleaving schemes} are developed to fully utilize the available per-objective evaluation budget \cite{allmendinger2015multiobjective}. More recently, the search experience of cheap objectives is leveraged to help and accelerate the optimization of expensive ones, thereby enhancing the overall efficiency in solving the problem. For example, Wang \emph{et al.} \cite{wang2021transfer} made use of domain adaptation techniques to align the solutions on/near the Pareto front in a latent space, which allows data augmentation for GPs of the expensive objectives. Alternatively, a co-surrogate model is introduced to capture the relationship between the cheap and expensive objectives in \cite{wang2022transfer}. Most recently, a new AF that takes both the search bias and the balance between exploration and exploitation into consideration was proposed \cite{wang2022heterogeneousMOPs}, thereby reducing the search bias caused by different per-objective evaluation times in multi/many-objective optimization.

Bayesian optimization for heterogeneous settings is still a new research field. This is particularly true when there are many expensive objectives but their computational complexities significantly differ. 

\subsection{Algorithmic fairness}
With the increasingly wider use of machine learning techniques in almost every field of science, technology and human life, there is a growing concern with the fairness of these algorithms. A large body of literature has demonstrated the necessity of avoiding discrimination and bias issues in finance, health care, hiring, and criminal justice that may result from the application of learning and optimization algorithms. A number of unfairness mitigation techniques have been dedicated to measuring and reducing bias/unfairness in different domains, which can be roughly divided into three groups, pre-, in-, and post processing, according to when the technique is applied \cite{perrone2021fair}. The first group aims to re-balance the data distribution before training the model. The second group typically trains the model either under fairness constraints or combining accuracy metrics with fairness, while the third group adjust the model after the training process. 

Accounting for fairness in the Bayesian optimization framework is a largely unexplored territory with few exceptions. For example, Perrone \emph{et al.} \cite{perrone2021fair} proposed an in-processing unfairness mitigation method in hyper-parameter optimization based on a constrained Bayesian optimization framework, called FairBO. In FairBO, an additional GP model is trained for the fairness constraint, allowing the constrained EI (cEI) to select new queries that satisfies the constraint. Unfortunately, such a constrained optimization method is designed for a single definition of fairness, which is not always applicable.  A different fairness concept was developed in a collaborative Bayesian optimization setting \cite{sim2021collaborative}, in which parties jointly optimize a black-box objective function. It is undesired for each collaborating party to receive unfair rewards while sharing their information with each other. Consequently, a new notion, called fair regret, is introduced based on fairness concepts from economics. Following the notion, the distributed batch GP-UCB is extended using a Gini social-evaluation function to balance the optimization efficiency and fairness.

The fairness problem in the context of Bayesian optimization is vital yet under-studied, and the measurement and mathematical definitions have not been explicit. Hence, the fairness definition should be well-defined at first, so that the fairness requirement can be more precisely integrated into the Bayesian optimization. The second fundamental open question is to investigate how fair surrogate models in Bayesian optimization are and how fair the selected new samples are. Finally, bias reduction strategies in Bayesian optimization can only be applied to the simplest case where a single fairness definition is adopted. The design of practical fairness-aware Bayesian optimization methods is still an open question.

\section{Conclusion}
\label{conclusion}
Bayesian optimization has become a popular and efficient approach to solving black-box optimization problems, and new methods have been emerging over the last few decades. In this paper, we performed a systematic literature review on Bayesian optimization, focused on new techniques for building the Gaussian process model and designing new acquisition functions to apply Bayesian optimization to various optimization scenarios. We divide these scenarios into nine categories according to the challenges in optimization, including high-dimensional decision and objective spaces, discontinuous search spaces, noise, constraints, and high computational complexity, as well as techniques for improving the efficiency of Bayesian optimization such as multi-task optimization, multi-fidelity optimization, knowledge transfer, and parallelization. Lastly, we summarize most recent developments in Bayesian optimization that address distributed data, data privacy, fairness in optimization, dynamism, and heterogeneity in the objective functions. So far, only sporadic research has been reported in these areas and many open questions remain to be explored. 

We hope that this survey paper can help the readers get a clear understanding of research landscape of Bayesian optimization, including its motivation, strengths and limitations, and as well as the future directions that are worth further research efforts.  

\bibliographystyle{ACM-Reference-Format}
\bibliography{sample-acmsmall}
\end{document}